\documentclass{article}

\usepackage{microtype}
\usepackage{graphicx}
\usepackage{multirow}
\usepackage{subcaption}
\usepackage{booktabs} % for professional tables

\usepackage{hyperref}

\usepackage[preprint]{icml2026}

\usepackage{mathtools}
\usepackage{amsthm}
\usepackage{amsmath}
\usepackage{amssymb}

\newcommand{\bx}{\mathbf{x}}
\newcommand{\bv}{\mathbf{v}}

\usepackage[capitalize,noabbrev]{cleveref}

\theoremstyle{plain}

\theoremstyle{definition}

\theoremstyle{remark}

\usepackage[disable,textsize=tiny]{todonotes}
\icmltitlerunning{Few-step Generative Models as Lossy Compression}

\begin{document}

\twocolumn[
  \icmltitle{Few-step Generative Models as Lossy Compression}


  \begin{icmlauthorlist}
    \icmlauthor{Fuma Kimishima}{hosei}
    \icmlauthor{Jinjia Zhou}{hosei}
  \end{icmlauthorlist}

  \icmlaffiliation{hosei}{Hosei University, Tokyo, Japan}

  \icmlcorrespondingauthor{Fuma Kimishima}{fuma.kimishima.3c@stu.hosei.ac.jp}
  \icmlcorrespondingauthor{Jinjia Zhou}{zhou@hosei.ac.jp}

  \icmlkeywords{Generative Models, Lossy Compression, Image Compression}

  \vskip 0.3in
]



\printAffiliationsAndNotice{}  % no special notice (required even if empty)

\begin{abstract}
DiffC provides a principled way to reuse pre-trained diffusion models for lossy compression, but its encoding and decoding procedures remain slow because they require many discretized forward and reverse steps. We study whether few-step generative models---Rectified Flow, Consistency Trajectory Models (CTM), and MeanFlow---can be cast as codecs within the same reverse channel coding (RCC) framework. The main challenge is that RCC requires posterior and shared distribution parameters, whereas these models do not explicitly parameterize intermediate conditional distributions. For Rectified Flow and MeanFlow, we use the equivalence between velocity parameterization and diffusion-style denoising parameterization to derive the quantities required by RCC. For CTM, which is distilled from EDM, we adopt the EDM noise parameterization together with local Gaussian approximations of the sender and shared distributions at intermediate states. This yields a proof-of-concept probabilistic formulation that enables compression with pre-trained few-step generative models without retraining. On low-resolution benchmarks, the resulting codecs reduce encoding and decoding time and improve realism in the low-bit-rate regime.
\end{abstract}    
\section{Introduction}
\label{sec:intro}
In lossy compression, \emph{realism} measures the distance between the distributions of original images and their reconstructions.
A realism value of zero indicates that reconstructions are statistically indistinguishable from the original data (\emph{perfect realism}).
Ho et al.~\cite{ho2020denoising} and Theis et al.~\cite{theis2022lossy} proposed a lossy compression method based on an unconditional diffusion model, the unconditional diffusion based codec, which is called DiffC.
DiffC interprets the forward (diffusion) process as an encoder and the reverse (denoising) process as a decoder: the encoder adds Gaussian noise, transmits the resulting noisy latent via reverse channel coding (RCC), and the decoder reconstructs the image through denoising or sampling.
In the idealized setting where samples generated by the diffusion model match the data distribution, DiffC can achieve perfectly realistic reconstructions.

In practice, DiffC supports multiple bit rates with a single pre-trained model, but its encoding and decoding procedures remain slow because they require many discretized diffusion steps to communicate Gaussian latents. The dominant bottleneck is decoding, where iterative denoising determines most of the runtime. This raises a natural question: can few-step generative models also be interpreted as codecs, thereby providing a faster alternative?
% Recent few-step generative models, reduce the cost of diffusion sampling by producing samples in one or a few steps. Representative examples include  
% Adapting such models to communication, however, requires posterior and shared distribution parameters for RCC, and these models do not explicitly parameterize the corresponding intermediate conditional distributions.

\begin{figure*}[t]
 \centering
 \includegraphics[width=\textwidth]{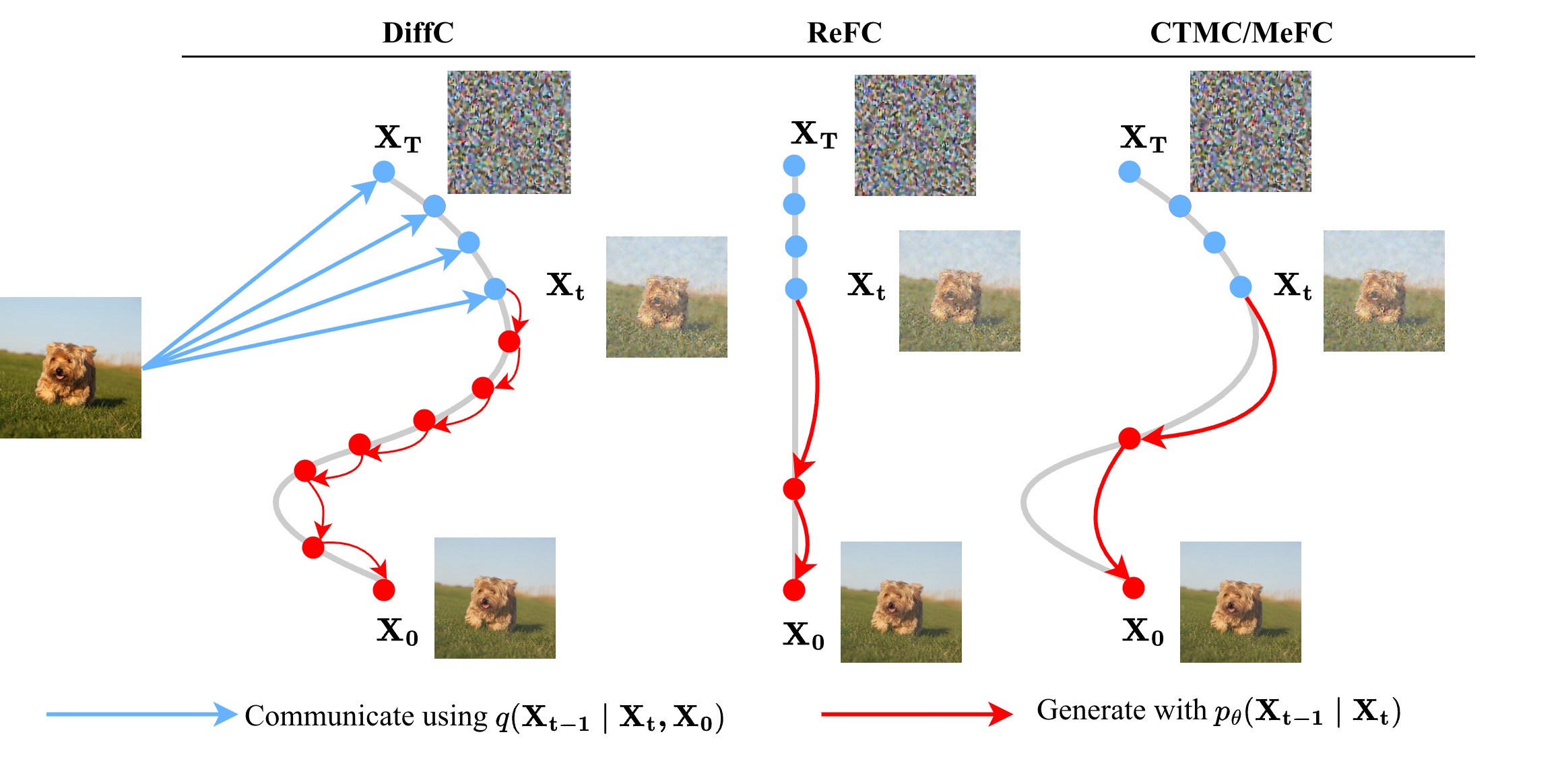}
 \caption{\label{fig:codec_overview} 
Overview of the reconstruction processes for different methods. The blue lines represent the communication steps while the red lines denote the generative process to reconstruct the image. Blue and red dots indicate the latent variables at specific timesteps $t$. The
gray lines illustrate the underlying trajectories learned by each generative model. While DiffC (left) requires multiple iterative steps to reconstruct an image, ReFC (center) and CTMC/MeFC (right) can generate the image much faster, either directly or by jumping along the trajectory.}
 \end{figure*}

In this work, we study how to treat few-step generative models, including Rectified Flow~\cite{liu2022flow}, Consistency Trajectory Models (CTM)~\cite{kim2023consistency}, and MeanFlow~\cite{geng2025mean}, as codecs in the same sense that diffusion models define the DiffC codec. The central technical problem is to obtain the posterior and shared distribution parameters required by RCC from models that were not designed to output them. For flow-based models, we derive these quantities through reparameterization, using the velocity--noise parameterization equivalence to rewrite the learned velocity field in diffusion-style denoising form. For CTM, we derive the required parameters by adopting the EDM noise parameterization and introducing local Gaussian approximations of the sender and shared distributions at intermediate states. This yields a codec construction for pre-trained few-step generative models without retraining. Across CIFAR10 and ImageNet $64\times64$, our works shows faster encoding and decoding together with stronger realism in the low-bit-rate regime.
Additional appendix results on ImageNet $256\times256$ further suggest that the proposed work retains similar qualitative advantages at higher resolution, producing higher fidelity reconstructions than current baselines in the low-bit-rate regime.

\section{Related Works}
\label{sec:Related Works}

%-------------------------------------------------------------------------
\subsection{Neural Image Compression}
Neural Image Compression (NIC) formulates lossy image compression as a rate-distortion optimization problem.
Most NIC methods adopt a variational autoencoder (VAE)~\cite{kingma2013auto}, in which the encoder, decoder, and entropy model are jointly trained~\cite{balle2016end, balle2018variational, theis2017lossy, agustsson2020universally}.
Blau and Michaeli~\cite{blau2018perception} characterized the trade-off between perceptual quality (realism) and reconstruction fidelity (distortion).
Building on this insight, many NIC approaches incorporate adversarial losses into VAE-based architectures~\cite{agustsson2019generative, galteri2017deep, mentzer2020high}, enabling explicit control over the distortion--realism trade-off.
More recently, conditional diffusion models have been used as decoders to reduce compression artifacts~\cite{hoogeboom2023high, yang2023lossy, careil2023towards, ohayon2025compressed}.
However, these approaches typically require training separate models for each target rate, limiting flexibility in practical deployments.

\begin{figure*}[t]
\begin{minipage}[t]{0.5\textwidth}
\begin{algorithm}[H]
 \caption{Sending $\bx_0$} \label{alg:sending}
 \small
 \begin{algorithmic}[1] 
    \STATE Send $\bx_T \sim q(\bx_T|\bx_0)$ using $p(\bx_T)$
    \FOR{$t=T$ to $2$}
     \STATE $\boldsymbol{\mu}_t \leftarrow \texttt{CalcMean\_Q}(\bx_t, \bx_0, \alpha_t, \sigma_t, \alpha_s, \sigma_s )$ % \gets -> \leftarrow
     \STATE $(\boldsymbol{\mu}_\theta, \hat \beta_t) \leftarrow \texttt{CalcMeanStd\_P}(\bx_t, \boldsymbol{\epsilon}_\theta, \alpha_t, \sigma_t, \alpha_s, \sigma_s)$
     \STATE $k_t = \frac{\boldsymbol{\mu}_t - \boldsymbol{\mu}_\theta}{\hat \beta_t}$
     \STATE Send $k_t$ using $p_\theta(\bx_{t-1} | \bx_{t})$
     \STATE $\bx_t = k_t \cdot \sigma_t + \boldsymbol{\mu}_\theta$
    \ENDFOR
    \STATE Send $\bx_0$ using $p_\theta(\bx_0|\bx_1)$
 \end{algorithmic}
\end{algorithm}
\end{minipage}
\hfill
\begin{minipage}[t]{0.5\textwidth}
\begin{algorithm}[H]
 \caption{Receiving} \label{alg:receiving}
 \small
 \begin{algorithmic}[1]
    \STATE Receive $\bx_T$ using $p(\bx_T)$ \COMMENT{Shared seed} % \triangleright -> \COMMENT
    \FOR{$t=T$ to $2$}
     \STATE $(\boldsymbol{\mu}_\theta, \bar \beta_t) \leftarrow \texttt{CalcMeanStd\_P}(\bx_t, \boldsymbol{\epsilon}_\theta, \alpha_t, \sigma_t, \alpha_s, \sigma_s)$
     \STATE Receive $k_t$ using $p_\theta(\bx_{t-1} | \bx_{t})$
     \STATE $\bx_t = k_t \cdot \bar \beta_t + \boldsymbol{\mu}_\theta$
     \STATE $\hat\bx_t = \bx_{\theta}(\bx_t; t)$
    \ENDFOR
    \STATE \textbf{return} $\hat\bx_0$
    \STATE \rule{0pt}{3.25ex} no operation
 \end{algorithmic}
\end{algorithm}
\end{minipage}
\vspace{-1em}
\end{figure*}

\subsection{Reverse Channel Coding}
To enable lossy compression with a pre-trained diffusion model, DiffC employs reverse channel coding (RCC) to communicate noisy latents.
For details of RCC, we refer the reader to prior work~\cite{li2018strong, theis2022algorithms, flamich2022fast}.
A primary challenge of RCC is its high computational complexity.

Recent research has proposed faster RCC variants~\cite{flamich2023faster, flamich2023greedy} to address this issue.
Vonderfecht et al.~\cite{vonderfecht2025lossy} introduced a practical RCC implementation that enables DiffC to handle high-resolution images within reasonable processing time.
Another promising direction is to use uniform noise instead of Gaussian noise, which can accelerate communication via universal quantization (UQ)~\cite{1057702, 119699}. Yang et al.~\cite{yang2024progressive} proposed UQDM, which injects uniform noise at an intermediate timestep and performs universal quantization, thereby enabling faster coding.
In this paper, we focus on constructing faster codecs from few-step generative models while retaining the RCC-based communication mechanism used in DiffC. Communication acceleration strategies that avoid RCC, such as UQ-based schemes with alternative posteriors at intermediate time, are left for future work.
\section{Preliminaries}

\subsection{Diffusion Models}
Denoising Diffusion Probabilistic Model (DDPM) models a data distribution via a latent-variable Markov chain.
Let $q(\bx_0)$ denote the data distribution, with observations $\bx_0\in\mathbb{R}^d$, and let $\bx_{1:T}\coloneqq (\bx_1,\ldots,\bx_T)$ denote latent variables.
The forward (diffusion) process $q(\bx_{1:T}|\bx_0)$ is defined as

\begin{align}
    \label{diffusion_posterior_q}
    q(\bx_{1:T}|\bx_0) &\coloneqq \prod_{t=1}^T q(\bx_t|\bx_{t-1})
    \\ q(\bx_t|\bx_{t-1})  &\coloneqq \mathcal{N}(\bx_t; \sqrt{1-\beta_t}\bx_{t-1}, \beta_t\mathbf{I})
\end{align}

where $\{\beta_t\}_{t=1}^T$ is the variance schedule, with $\beta_t\in \mathbb{R}$ at time $t$, and is typically designed so that $\bx_t$ approaches a sample from $\mathcal{N}(0, \mathbf{I})$. In other words, the forward process gradually adds Gaussian noise to the data, transforming it into a standard normal distribution. To generate samples, DDPM inverts this process through the reverse process $p_\theta(\bx_{0:T})$, which starts from $p(\bx_T)=\mathcal{N}(0, \mathbf{I})$:

\begin{align}
    \label{diffusion_posteriro_p}
    p_\theta(\bx_{0:T}) &\coloneqq \prod_{t=1}^T p_\theta(\bx_{t-1}|\bx_t) \\
    \\ p_\theta(\bx_{t-1}|\bx_t) &\coloneqq \mathcal{N}(\bx_{t-1}; \boldsymbol{\mu}_\theta(\bx_t,t), \boldsymbol{\Sigma}_\theta(\bx_t, t)\mathbf{I})
\end{align}

where $\boldsymbol{\mu}_\theta(\bx_t,t)$ and $\boldsymbol{\Sigma}_\theta(\bx_t, t)$ are estimation targets, and details are provided in Appendix. Here, networks are optimized by minimizing the negative evidence log likelihood (NELBO) below: 

\begin{equation}
\label{eq:ELBO}
\begin{aligned}
\mathcal{L}(\bx_0)
= \mathbb{E}_q\Big[
& D_{\mathrm{KL}}\bigl(q(\bx_T\mid\bx_0)\,\Vert\,p(\bx_T)\bigr) \\
& + \sum_{t=2}^{T} D_{\mathrm{KL}}\bigl(q(\bx_{t-1}\mid\bx_t,\bx_0)\,\Vert\,p_\theta(\bx_{t-1}\mid\bx_t)\bigr) \\
& - \log p_\theta(\bx_0\mid\bx_1)
\Big].
\end{aligned}
\end{equation}

Throughout the paper, logarithms are in base 2.
The reverse process is trained to match the forward-process posteriors at each time step.

To simplify the notation in later sections, we parameterize the noise schedule using two scalar functions of time, $\alpha(t)$ and $\sigma(t)$, which control the signal and noise scales at time $t$.
We write $\alpha_t\coloneqq\alpha(t)$ and $\sigma_t\coloneqq\sigma(t)$.
The schedule is chosen such that $\bx_T$ is close to a standard normal distribution.
With this parameterization, a sample from the forward process can be written as

\begin{align}
\label{forward_diffusion_sample}
    \bx_t &= \alpha_t\bx_0 + \sigma_t\boldsymbol{\epsilon},  \quad \boldsymbol{\epsilon} \sim\mathcal{N}(0, I)
\end{align}

The reverse process is defined by the same transformation equations as the forward process, and estimates samples at a lower noise level $s$ from $t$ with two estimators: $\boldsymbol{\epsilon}_\theta = \boldsymbol{\epsilon}_\theta(\bx_{t}, t)$ that predicts the added noise at time $t$ (a.k.a noise-prediction model), and $\bx_\theta=\bx_\theta(\bx_{t}, t)$ which predicts the plausible $\bx_0$ from $\bx_{t}$ (a.k.a denoising model):

\begin{align}
\label{reverse_diffusion_sample}
    \bx_s &= \alpha_t\bx_{\theta} + \sigma_t\boldsymbol{\epsilon}_{\theta} \\
\label{x0_estimation}
    \bx_\theta &= \frac{\bx_{t}-\sigma_t\boldsymbol{\epsilon}_\theta}{\alpha_t}
\end{align}

\subsection{DiffC}
DiffC~\cite{theis2022lossy} is a family of lossy compression algorithms built on unconditional diffusion models.
It performs progressive compression by gradually adding Gaussian noise and communicating intermediate noisy latents via reverse channel coding (RCC).

To compress $\bx_0$, the sender can sample from the true posterior $q(\bx_t\mid\bx_0)$ (the forward process), whereas the receiver cannot access $\bx_0$ and instead relies on the shared distribution $p_\theta(\bx_{t-1}\mid\bx_t)$.
Under RCC, the sender and receiver share the random seed.
At the initial step $t=T$, the sender transmits a sample $\bx_T\sim q(\bx_T\mid\bx_0)$ using $p(\bx_T)$.
Then, for $t=T,\ldots,2$, the sender transmits information that enables the receiver to recover $\bx_{t-1}\sim q(\bx_{t-1}\mid\bx_t,\bx_0)$ using $p_\theta(\bx_{t-1}\mid\bx_t)$.
Finally, at $t=0$, $\bx_0$ is losslessly compressed using $p_\theta(\bx_0\mid\bx_1)$.

It is known that if the encoder and decoder share a source of randomness (typically implemented with a pseudo-random number generator) in RCC, the coding cost of $\bx \sim q$ using $p$ is equal to the KL divergence, $D_{KL}(q \| p)$, in bits~\cite{theis2022algorithms}. Therefore, for DiffC, the coding cost is naturally related to the NELBO defined in Eq.~\ref{eq:ELBO}. As noted by~\cite{ho2020denoising}, the NELBO $\mathcal{L}(\bx_0)$ satisfies $\mathcal{L}(\bx_0)\ge \mathbb{E}_q \left[-\log p_\theta(\bx_0) \right]$, meaning that DDPMs optimize a variational upper bound on the negative log likelihood, which is a standard objective for data compression. In other words, given a pre-trained DDPM, performing the lossy compression procedure described above achieves the coding cost characterized by the NELBO. This is why using a well-trained diffusion model (e.g., Stable Diffusion) improves rate--distortion performance~\cite{vonderfecht2025lossy}.

After the receiver obtains $\bx_t$, the reconstruction $\hat \bx_0$ can be obtained using the denoising model $\bx_\theta$, ancestral sampling, or a probability flow ODE~\cite{ho2020denoising, theis2022lossy}. Assuming the diffusion model perfectly captures the data distribution, the reconstructed sample $\hat \bx_0$ is indistinguishable from the original data; this approach therefore achieves perfect realism in theory~\cite{theis2022lossy}.

Although DiffC provides a principled codec based on diffusion models, its reliance on many iterative denoising steps introduces a major computational bottleneck. This motivates the following question: can few-step generative models also be treated as codecs and thereby provide a faster alternative to DiffC? In this paper, we answer this question by deriving the posterior and shared distribution parameters required by RCC for Rectified Flow, Consistency Trajectory Models (CTM), and MeanFlow.
\section{Method}

\subsection{Detail in DiffC Algorithm}
Vonderfecht et al.~\cite{vonderfecht2025lossy} implemented DiffC with practical runtime, as summarized in Algorithms~\ref{alg:sending} and~\ref{alg:receiving}.
In each RCC step, DiffC computes the mean (and variance) of the sender distribution $q$ and the shared distribution $p$ via $\texttt{CalcMean\_Q}$ and $\texttt{CalcMeanStd\_P}$ which are given in Alg.~\ref{alg:calc_mean_q} and \ref{alg:calc_mean_std_p}.
To communicate the sample efficiently, Vonderfecht et al.~\cite{vonderfecht2025lossy} transmit the standardized difference
$\boldsymbol{k}_t = (\boldsymbol{\mu}_t-\boldsymbol{\mu}_\theta)/\sigma_t$ (see Appendix A.4 of~\cite{vonderfecht2025lossy}).
After $\boldsymbol{k}_t$ is received, the receiver can recover $\bx_t$ and reconstruct the final image using the pre-trained denoiser or sampler.

In DDPM-based DiffC, the means of both $q$ and $p$ can be computed analytically.
In contrast, for Rectified Flow, CTM, and MeanFlow, the intermediate-time distributions are not available in closed form, and these models do not necessarily define an explicit forward diffusion process.
They instead learn transport trajectories (or jumps along them) without explicitly modeling the marginal distribution at each time $t$.
To instantiate codecs from such few-step generative models, we derive model-specific versions of $\texttt{CalcMean\_Q}$ and $\texttt{CalcMeanStd\_P}$.

\subsection{Rectified Flow as a Codec}
Rectified Flow learns a deterministic transport map that connects two distributions with a straight path. This design simplifies the solution of ODE, eliminating the need for iterative numerical integration and resulting in faster computation. Formally, given data $\bx_1 \sim p_1$ and prior $\bx_0\sim p_0=\mathcal{N}(0, I)$ (note that we adjust the notation in Rectified Flow paper~\cite{liu2022flow} which is the reverse notation commonly used in diffusion literature), the flow at time $t$ is defined as follows:

\begin{equation}
\label{RF_interpolation}
    \bx_t = t\bx_1+(1-t)\bx_0
\end{equation}

This can be regarded as a specific instance where $\alpha(t)=t=\alpha_\text{RF}(t)$ and $\sigma(t)=1-t=\sigma_\text{RF}(t)$ in Eq.~\ref{forward_diffusion_sample}.
The velocity $\bv_t$ is naturally derived as $\bv_t = \frac{d \bx_t}{d t}=\bx_1 - \bx_0$. Then, a neural network $\bv_\theta^\text{RF}(\bx_t, t)$ is trained to approximate the ideal straight path $\bv_t$ by minimizing the following loss function:

\begin{equation}
\label{RF_objective}
    \mathcal{L}(\theta)
    = \mathbb{E}_{\bx_0, \bx_1, t}
      \big[ \| \bv_\theta^\text{RF}(\bx_t, t) - (\bx_1 - \bx_0) \|^2 \big].
\end{equation}

The trajectory is progressively straightened by recursively applying the rectification process. As a result, it can be simulated accurately using only a few discretization steps. To generate a new sample, the ODE $\left(\frac{d \bx_t}{d t} = \bv_\theta^\text{RF}(\bx_t, t)\right)$ is solved for $\bx_t$.

Algorithms~\ref{alg:sending} and~\ref{alg:receiving} require a noise-prediction model $\boldsymbol{\epsilon}_\theta$ inside $\texttt{CalcMeanStd\_P}$ although Rectified Flow does not directly provide it.
Prior work~\cite{vonderfecht2025lossy} aligned the Rectified Flow noise schedule with that of diffusion (called DiffC (RF) in this paper. See detail in Appendix~\ref{app:diffc-rf}), but still relied on an indirect translation.
Our objective is to derive codec parameters directly from $\bv_\theta^\text{RF}$. To do so, we express Rectified Flow in the $\boldsymbol{\epsilon}$-prediction form leveraging the velocity--noise correspondence characterized by Lai et al.~\cite{lai2025principles}. For Rectified Flow, denoting the resulting predictor by $\boldsymbol{\epsilon}_\theta^{\text{RF}}$, we obtain:

\begin{equation}
\label{RF_velocity_to_epsilon}
    \boldsymbol{\epsilon}_{\theta}^\text{RF}(\bx_t, t) = \frac{\bv_{\theta}^\text{RF}(\bx_t,t)-\frac{\alpha_\text{RF}(t)'}{\alpha_\text{RF}(t)}\bx_t}{\sigma_\text{RF}(t)'-\frac{\alpha_\text{RF}(t)'\sigma_\text{RF}(t)}{\alpha_\text{RF}(t)}}
\end{equation}

The Appendix specializes this general equivalence to our notation. Substituting this expression into Eq.~\ref{x0_estimation} yields the parameters used to define \emph{Rectified Flow as a Codec (ReFC)}.
As visualized in Fig.~\ref{fig:codec_overview}, ReFC follows a straight path during decoding, which leads to efficient reconstruction.

\subsection{CTM as a Codec}
CTM accelerates diffusion sampling by learning \emph{anytime-to-anytime} jumps along the probability flow ODE (PF-ODE) trajectory~\cite{kim2023consistency}.
This property enables generation with only a few steps, or even in one step.
In this setting, the trajectory spans from $\bx_0$ (data) to $\bx_1$ (Gaussian noise).
CTM defines a predictor $\boldsymbol{G}_{\theta}(\bx_t, t, s)$ as the PF-ODE solution mapping a sample from time $t$ to time $s$, where $0 \leq s \leq t \leq 1$:

\begin{equation}
    \boldsymbol{G}_\theta(\bx_t, t, s) := \frac{s}{t} \bx_t + \left(1 - \frac{s}{t}\right) \boldsymbol{g}_\theta(\bx_t, t, s)
\end{equation}

where $\boldsymbol{g}_\theta(\bx_t, t, s)=\bx_t + \frac{t}{t-s}\int_t^s \frac{\bx_u - E[\bx|\bx_u]}{u} du$, which approximates the $t$-to-$s$ jump when $t \ne s$ and the denoising function when $t=s$.

Unlike DDPM, CTM does not define an explicit forward process.
To cast CTM as a codec, we introduce the following intermediate-time distributions at time $t$:

\begin{align}
\label{CTM_distribution}
    q_t^\text{CTM} &= \mathcal{N}(\alpha_\text{CTM}(t)\bx_0, \sigma_\text{CTM}(t)^2\boldsymbol{I}) \\
    p_t^\text{CTM} &= \mathcal{N}(\alpha_\text{CTM}(t)\bx_\theta^\text{CTM}, \sigma_\text{CTM}(t)^2\boldsymbol{I})
\end{align}

where $\alpha_{\text{CTM}}(t)$ and $\sigma_{\text{CTM}}(t)$ denote the noise-schedule parameters.
Because the CTM used in our experiments is distilled from EDM~\cite{karras2022elucidating}, we follow the EDM parameterization with $\alpha_{\text{CTM}}(t)=1$ and

\begin{equation}
\label{eq:edm_sigma_schedule}
\sigma_{\text{CTM}}(t)
= \left(\sigma_{\max}^{1/\rho} + t\left(\sigma_{\min}^{1/\rho}-\sigma_{\max}^{1/\rho}\right)\right)^{\rho}, \quad t\in[0,1]
\end{equation}

where $\sigma_{\min}>0$, $\sigma_{\max}>\sigma_{\min}$, and $\rho>0$ follow~\cite{karras2022elucidating}.
$\bx_\theta^\text{CTM}$ is a function that predicts the clean data $\bx_0$, where we employ $\boldsymbol{g}_\theta(\bx_t, t, t)$ as this function.
These densities $q_t^\text{CTM}$ and $p_t^\text{CTM}$ are centered around $\bx_t$ and describe how $\bx$ is perturbed when Gaussian noise with scale $\sigma_t$ is added. Although the ELBO directly corresponds to the coding cost in DiffC, quantifying the exact coding cost induced by these assumed distributions remains an open problem and is left for future work. In this paper, we instead focus on an empirical evaluation of the achievable rate--distortion performance under this formulation.

For a pre-trained CTM,
we define $\texttt{CalcMean\_Q}$ and $\texttt{CalcMeanStd\_P}$ so that they compute the mean of $q_t^{\text{CTM}}$ and the mean/variance of $p_t^{\text{CTM}}$, respectively.
These quantities follow directly from Eq.~\ref{CTM_distribution}, and we refer to the resulting construction as \emph{CTM as a Codec (CTMC)}.
On the receiver side, we use $\boldsymbol{g}_\theta(\bx_t,t,s)$ to jump along the trajectory for faster reconstruction, as illustrated in Fig.~\ref{fig:codec_overview}.

\begin{figure*}[t]
 \centering
 \includegraphics[width=\textwidth]{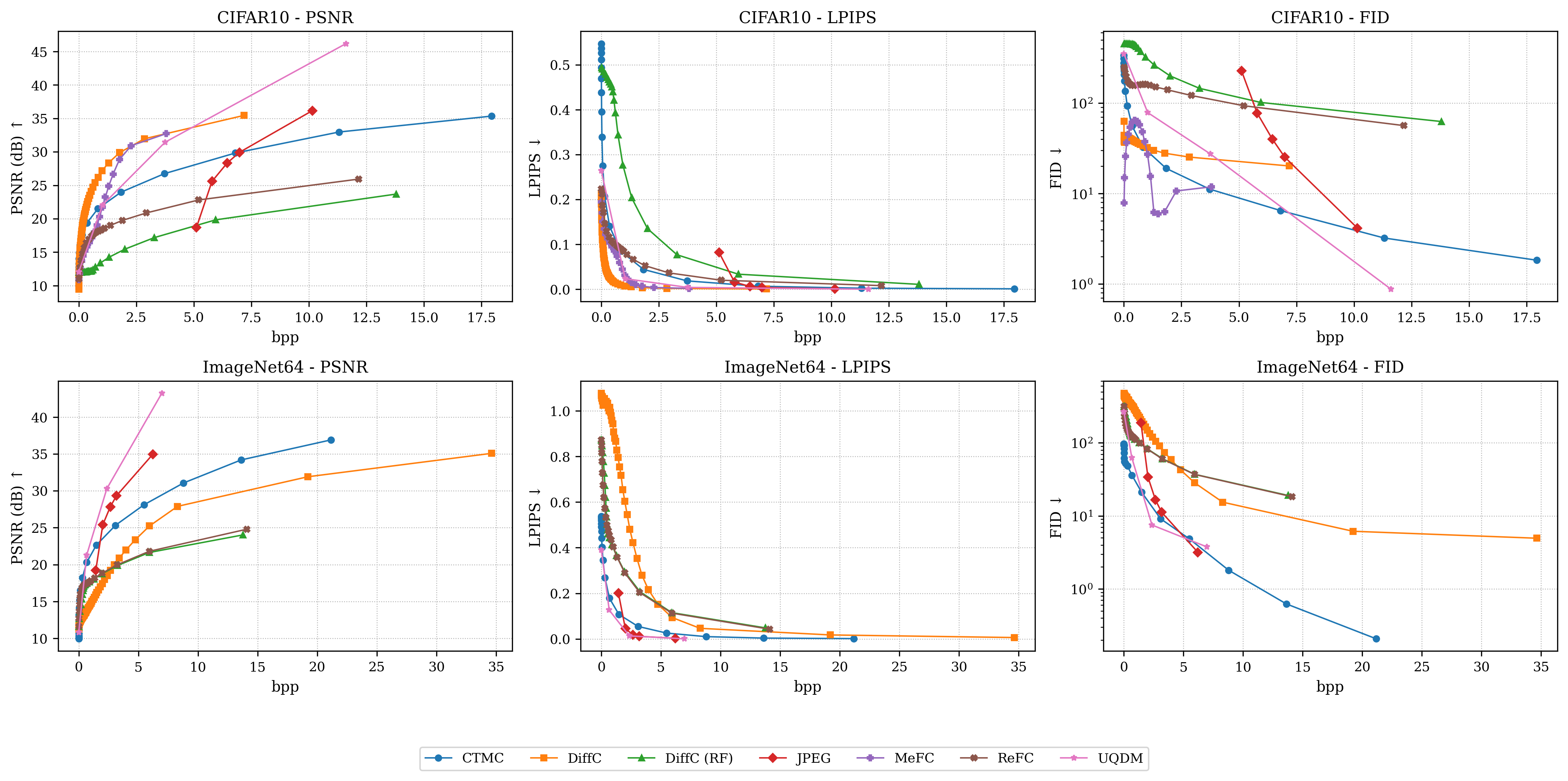}
 \caption{\label{fig:bpp_metric_plot} Rate–distortion and rate–realism curves, with distortion (PSNR and LPIPS) and realism (FID), plotted against bits per pixel (bpp). The top row shows results on CIFAR10, and the bottom row shows results on ImageNet $64\times64$.
}
 \end{figure*}

\subsection{MeanFlow as a Codec}
Rectified Flow learns an instantaneous velocity field, but the resulting transport trajectory is not guaranteed to be perfectly straight in practice.
Its objective in Eq.~\ref{RF_objective} averages over the data distribution and does not connect data points with a straight path. Furthermore, the original paper empirically reports that repeatedly applying the reflow process accumulates error~\cite{liu2022flow}. This indicates that the resulting paths may still be curved and may therefore require multiple discretization steps for high-quality generation.
MeanFlow addresses this limitation by learning an average velocity field that represents the displacement between two time points $t$ and $s$ ($s\leq t$)~\cite{geng2025mean}.
The average velocity $\boldsymbol{u}$ is defined as

\begin{equation}
    \boldsymbol{u}(\bx_t,t,s) \triangleq \frac{1}{s-t}\int_t^s \boldsymbol{v}^{\text{MF}}(\bx_\gamma, \gamma)d \gamma
\end{equation}

where $\boldsymbol{v^\text{MF}}(\bx_t,t)$ is the velocity field which coincides with the conditional flow matching (CFM)~\cite{lipman2022flow}.
The neural network $\boldsymbol{u}_{\theta}(\bx_t, t, s)$ is trained to fit the target $\boldsymbol{u}_\text{tgt}$ which is defined as:

\begin{equation}
    \boldsymbol{u}_\text{tgt} = \boldsymbol{v^\text{MF}}(\bx_t, t) - (s-t)(\boldsymbol{v^\text{MF}}(\bx_t, t)\partial_\bx \boldsymbol{u}_\theta + \partial_t \boldsymbol{u}_\theta)
\end{equation}

When $t=s$, the learned field $\boldsymbol{u}_\theta(\bx_t,t,t)$ coincides with the instantaneous velocity $\boldsymbol{v}^{\text{MF}}(\bx_t,t)$.
This objective can be expressed in a form analogous to Eq.~\ref{RF_objective}, so the codec parameters can again be derived from the learned flow field.
Specifically, with $\alpha_{\text{MF}}(t)=t$ and $\sigma_{\text{MF}}(t)=1-t$, we express MeanFlow in the noise-prediction, leveraging the velocity--noise parametrization ~\cite{lai2025principles}. As same with ReFC, this yields the following predictor $\boldsymbol{\epsilon}_\theta^{\text{MF}}$:

\begin{equation}
\label{MF_velocity_to_epsilon}
    \boldsymbol{\epsilon}_{\theta}^\text{MF}(\bx_t, t) = \frac{\boldsymbol{u}_{\theta}^\text{MF}(\bx_t,t,t)-\frac{\alpha_\text{MF}(t)'}{\alpha_\text{MF}(t)}\bx_t}{\sigma_\text{MF}(t)'-\frac{\alpha_\text{MF}(t)'\sigma_\text{MF}(t)}{\alpha_\text{MF}(t)}}
\end{equation}

We use $\boldsymbol{\epsilon}_{\theta}^{\text{MF}}(\bx_t,t)$ in $\texttt{CalcMeanStd\_P}$ to obtain the parameters required for encoding and decoding.
This defines \emph{MeanFlow as a Codec (MeFC)}.
MeanFlow shares a key property with CTM---the ability to jump between arbitrary time points---which enables efficient reconstruction by shortcutting the trajectory.

\section{Experiments}

\begin{table}[t]
\caption{The number of steps and average encoding/decoding time (in seconds) comparison tested on CIFAR10. "Fwd" and "Rev" refer to forward and reverse process, respectively.}
\setlength{\tabcolsep}{1.5mm}
\label{table:CIFAR10}
\begin{tabular}{cccccc}
\cline{1-5}
Method & Fwd Steps & Rev Steps & Enc (s) & Dec (s) &  \\
\cline{1-5}
DiffC   & 50 & 50     &   2.95      &   48.26  &  \\
DiffC (RF) & 20 & 20 & 1.62 & 4.38 \\
UQDM  & 4 & 4  &    0.50    &   0.31    &  \\
ReFC  & 20 & 20   &  1.61   &   4.24  &  \\
CTMC  & 20 & 1   & 3.71    &   0.93    &  \\
MeFC  & 20 &  20  &  0.93    &   4.95   &  \\
\cline{1-5}
\end{tabular}
\end{table}

\begin{figure*}
 \centering
 \includegraphics[width=\textwidth]{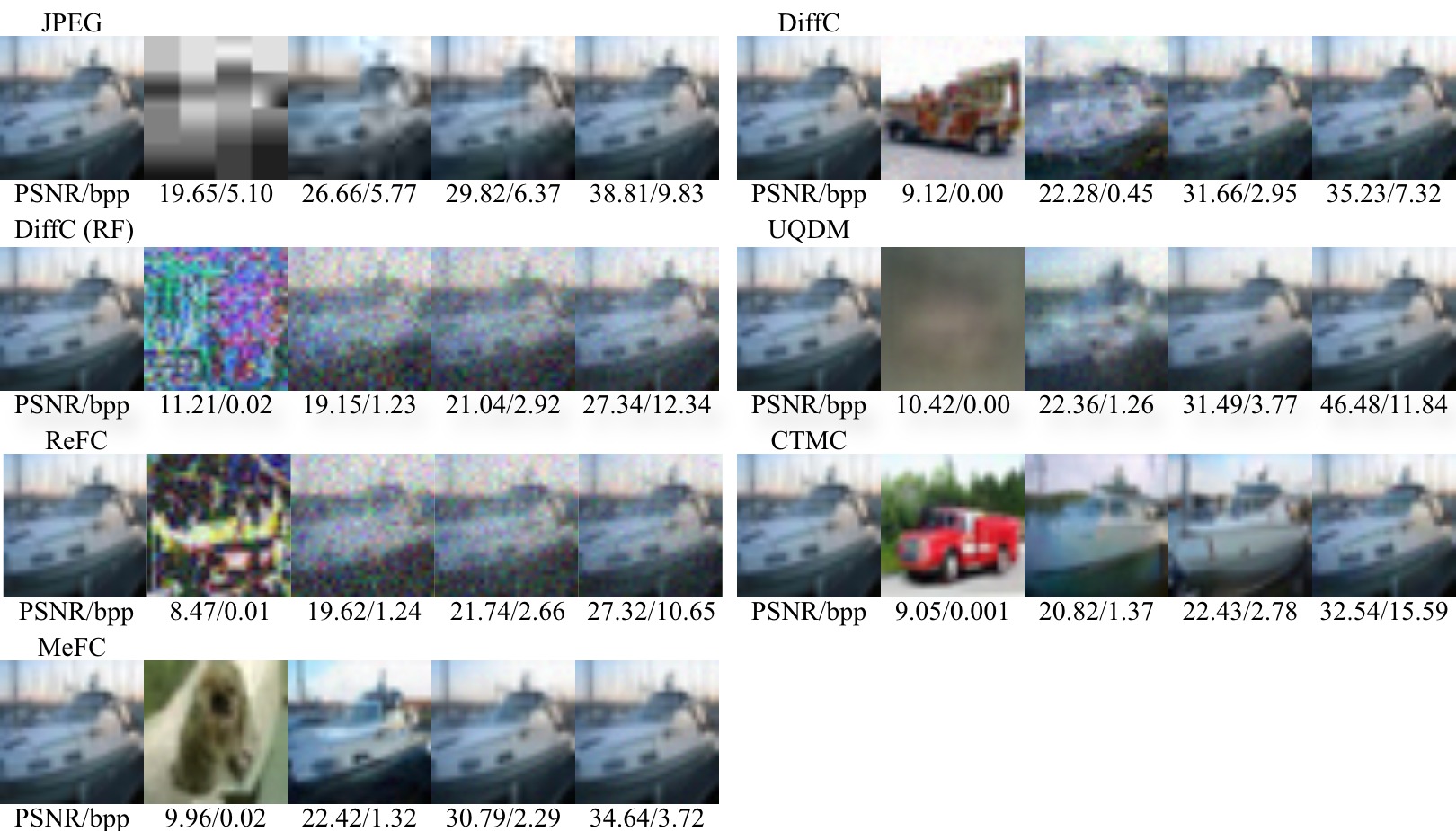}
 \caption{\label{fig:CIFAR10_visual_comparison} Reconstructed images on the CIFAR10 at various PSNR and bpp. For DiffC, UQDM, CTMC, and MeFC, PSNR/bpp values are selected to enable a comparative analysis of their performance evaluation. In contrast, DiffC (RF) and ReFC are chosen from approximately the same bpp for direct comparison.}
 \end{figure*}

 \begin{figure*}[t]
 \centering
 \includegraphics[width=\textwidth]{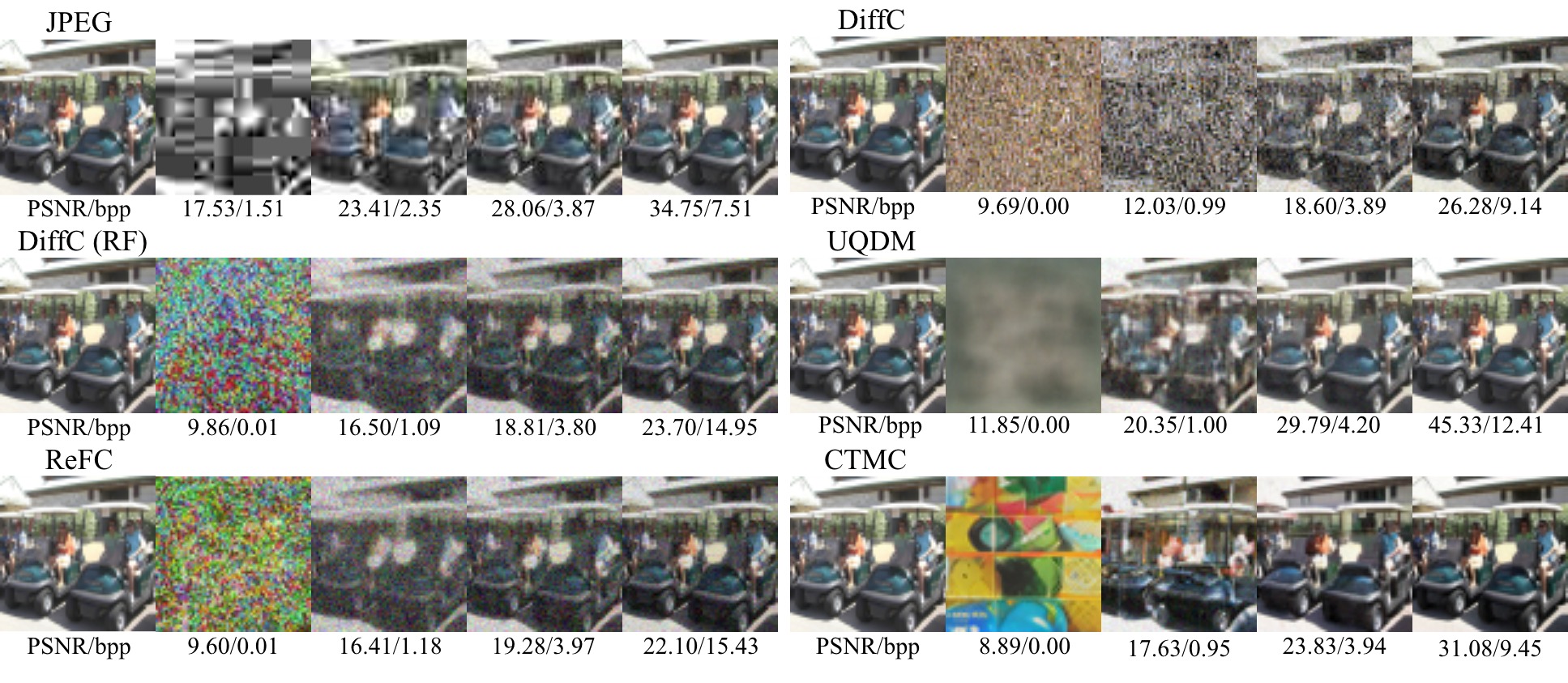}
 \caption{\label{fig:imagenet64_visual_comparison} Reconstructed images on ImageNet $64\times64$ dataset at various PSNR/bpp. Results are approximately aligned to match the bpp of UQDM.
}
 \end{figure*}

\subsection{Settings}
We evaluate the proposed codecs on two low-resolution datasets; CIFAR10~\cite{krizhevsky2009learning} and ImageNet $64\times64$~\cite{5206848, imagenet15russakovsky} as a proof of concept.
Additional results on ImageNet $256\times256$ are reported in the Appendix.
For ReFC, CTMC, and MeFC, we use publicly available pre-trained models~\cite{lee2024improving, kim2023consistency, geng2025mean, meanflow_pytorch} summarized in Table~\ref{table:repositories}.
We compare against the traditional codec JPEG, the diffusion-based progressive codec DiffC~\cite{theis2022lossy}, DiffC (RF), and the faster variant UQDM~\cite{yang2024progressive}. 
% DiffC (RF) is the Rectified-Flow-based codec derived by Vonderfecht et al.~\cite{vonderfecht2025lossy}, which incorporates Rectified Flow into the DiffC algorithm as described in Appendix~\ref{}.

\subsection{CIFAR10}
We first show the configurations of each method used in the CIFAR10 experiments and a encoding and decoding time comparisons in Table~\ref{table:CIFAR10}.
We fix the number of forward steps to 20 for ReFC, CTMC, and MeFC, and set the reverse steps so that each method could obtain high quality reconstructions with shorter processsing time; additional results with fewer reverse steps (including one-step reconstruction) are provided in the Appendix.
We measure the per-image runtime by encoding and decoding 10 images individually on a single NVIDIA RTX~3090 GPU.
Compared to DiffC, the proposed codecs achieve similar encoding times and lower decoding times across the evaluated bit rates.
Among the baselines, UQDM achieves the fastest encoding and decoding.
We attribute this to its use of universal quantization and to its four-step diffusion backbone~\cite{yang2024progressive}.
CTMC, ReFC, and MeFC currently use RCC; combining few-step codecs with universal quantization is a natural direction for further speedups.

Fig.~\ref{fig:bpp_metric_plot} (top) reports rate-distortion and rate-realism curves, plotting bits per pixel (bpp) against distortion (PSNR and LPIPS~\cite{zhang2018unreasonable}) and realism (FID~\cite{heusel2017gans}).
These results show that the codec construction from few-step generative models preserves the main advantage of DiffC while reducing runtime. MeFC attains the best distortion performance at a given bpp in terms of both PSNR and LPIPS on CIFAR10. CTMC reduce distorition further in the low-bit-rate regime, although it requires higher bpp to match the PSNR and LPIPS of the best-performing methods at higher rates. ReFC improves over DiffC (RF), but remains weaker than CTMC and MeFC. In the rate-realism curves, CTMC and MeFC achieve better realism than DiffC and UQDM at fixed bpp, whereas ReFC shows a flatter improvement trend.

MeFC exhibits a non-monotonic rate--realism trend on CIFAR10: while FID typically decreases as bpp increases, the MeFC curve shows noticeable fluctuations. This behavior can be empirically explained by the MeanFlow trajectory. 
As shown in Fig.~\ref{fig:mefc_steps_cifar10}, when reconstructing from latents at small timesteps ($t=0,1,2,3$), MeanFlow produces perceptually natural images even under lossy compression settings, indicating that it can generate realistic samples from Gaussian noise. In contrast, for intermediate timesteps ($t=4,5,6,7,8$), the denoising process remains incomplete and the reconstructed images retain noticeable noise. As $t$ further increases, the residual noise is gradually removed and the reconstructions become progressively cleaner.

%When reconstructing from latents at small $t$, MeanFlow tends to generate samples that match the natural image distribution. In contrast, for latents at larger $t$, it reconstructs the input with added Gaussian noise (Fig.~\ref{fig:mefc_steps_cifar10}).

Fig.~\ref{fig:CIFAR10_visual_comparison} provides qualitative comparisons.
At low bit rates, while CTMC and MeFC produce reconstruction results that look different from the input images but still resemble natural scenes (such as a red car or a sitting dog), UQDM outputs grayscale images. 
The ability of CTM and MeanFlow to generate natural images from Gaussian noise in just a few steps is useful in lossy compression settings, as it allows for the restoration of natural images at low bit rates. In contrast, UQDM injects uniform noise at intermediate timesteps and learns reverse kernels that are convenient for coding via universal quantization. Although its forward process is designed so that the terminal distribution converges to a Gaussian in the limit, this does not imply that the learned reverse process can reliably generate natural images from a Gaussian prior. We therefore hypothesize that this mismatch is one reason why UQDM struggles to produce realistic reconstructions at low bitrates.
ReFC remains noisier overall but is typically less noisy than DiffC (RF) at low bit rate.
We expect stronger pre-trained Rectified Flow backbone to further improve the denoising ability.

\subsection{ImageNet 64$\times$64}

\begin{table}[t]
\caption{Comparison of the number of steps for both forward ("Fwd") and reverse ("Rev"), and average encoding (Enc) and decoding (Dec) times in seconds. All timings were measured on a single RTX 3090 GPU for ImageNet 64$\times$64.}
\setlength{\tabcolsep}{1.5mm}
\label{table:ImageNet 64x64}
\begin{tabular}{cccccc}
\cline{1-5}
Method & Fwd Steps & Rev Steps & Enc (s) & Dec (s) &  \\
\cline{1-5}
DiffC  & 50 & 50     &       19.92     & 47.48 &  \\
DiffC (RF)  & 20 & 20     &    5.83    & 7.74 &\\
UQDM  & 4 & 4  &   0.80    &   0.57    &  \\
ReFC & 20 & 20   &   5.67   &   7.95  &  \\
CTMC   & 20 &  1   &   9.35     &     1.25    &  \\
\cline{1-5}
\end{tabular}
\end{table}

Table~\ref{table:ImageNet 64x64} summarizes the numbers of forward and reverse process steps, together with the encoding and decoding times, for each method evaluated on ImageNet $64\times64$.
We report the average time required to encode and decode a single image, measured by processing 10 images on a single NVIDIA RTX~3090 GPU. For ReFC and CTMC, the number of steps is chosen to balance reconstruction quality and processing time.
In particular, the CTMC reverse process is configured as one-step.

The observed trends are similar to those on CIFAR10.
Both ReFC and CTMC reduce encoding and decoding time relative to DiffC, while remaining slower than UQDM.
The rate--distortion and rate--realism curves on ImageNet $64\times64$, which are shown in Fig.~\ref{fig:bpp_metric_plot} (bottom), support the proposed codec construction from few-step generative models; CTMC and UQDM achieve comparable PSNR in the low-bit-rate regime, but CTMC attains better realism. As the bit rate increases, the PSNR advantage of CTMC becomes smaller. The gap between ReFC and DiffC (RF) is smaller than that on CIFAR10 in both distortion and realism. These trends are consistent with the visual comparisons in Fig.~\ref{fig:imagenet64_visual_comparison}. At low bit rates, CTMC produces more plausible reconstructions, whereas ReFC and DiffC (RF) remain noisier; across additional samples, ReFC is generally less noisy than DiffC (RF).

\section{Discussion}
In this paper, we studied how pre-trained few-step generative models such as Rectified Flow, CTM, and MeanFlow can be interpreted as codecs (ReFC, CTMC, and MeFC).
A key step for flow-based compression is to use the velocity--noise parameterization to express learned velocity fields in $\boldsymbol{\epsilon}$-prediction form, thereby obtaining the parameters required to define the codec.
Compared to prior flow-based adaptations that rely on aligning noise schedules, this direct conversion offers a cleaner probabilistic route for constructing codecs from flow-based generative models.
Taken together, the experiments suggest that few-step reconstruction---including one-step decoding with CTMC---offers a favorable trade-off between runtime and realism relative to DiffC on the low-resolution benchmarks considered here.
Two natural directions for future work are: (i) incorporating high-resolution generative backbones, including models such as Flux, to test whether these constructions extend beyond low-resolution benchmarks, and (ii) designing few-step generative models with uniform noise perturbations as UQDM, and combining them with universal quantization to further reduce coding time.

\bibliography{example_paper}
\bibliographystyle{icml2026}

%%%%%%%%%%%%%%%%%%%%%%%%%%%%%%%%%%%%%%%%%%%%%%%%%%%%%%%%%%%%%%%%%%%%%%%%%%%%%%%
%%%%%%%%%%%%%%%%%%%%%%%%%%%%%%%%%%%%%%%%%%%%%%%%%%%%%%%%%%%%%%%%%%%%%%%%%%%%%%%
% APPENDIX
%%%%%%%%%%%%%%%%%%%%%%%%%%%%%%%%%%%%%%%%%%%%%%%%%%%%%%%%%%%%%%%%%%%%%%%%%%%%%%%
%%%%%%%%%%%%%%%%%%%%%%%%%%%%%%%%%%%%%%%%%%%%%%%%%%%%%%%%%%%%%%%%%%%%%%%%%%%%%%%
\newpage
\appendix
\onecolumn

\section{DDPM}
As a reference and to explain the two functions "$\texttt{CalcMeanStd\_Q}$" and "$\texttt{CalcMeanStd\_P}$" in Algorithm~\ref{alg:sending} and \ref{alg:receiving}, we provide a brief review of DDPM~\cite{ho2020denoising}. 

As described in the main paper, DDPM optimize the ELBO in Eq.~\ref{eq:ELBO}, where the forward process posteriors $q(\bx_{t}|\bx_{t+1}, \bx_0)$ has the closed formulation as follow:

\begin{align}
    \label{eq:posterior}
    q(\bx_{t-1}|\bx_{t}, \bx_0) &= \mathcal{N}(\bx_{t-1}; \boldsymbol{\tilde \mu}(\bx_{t}, \bx_0), \tilde \beta_{t}\mathbf{I}) \\
    \label{eq:mu_forward_closed_form}
    \boldsymbol{\tilde \mu}(\bx_{t}, \bx_0) &\coloneqq  \frac{\sqrt{\bar\alpha_{t-1}^{\text{DDPM}}} \beta_{t}}{1-\bar \alpha_{t}} \bx_0 + \frac{\sqrt{\alpha_{t}^{\text{DDPM}}}(1-\bar \alpha_{t+1})}{1-\bar \alpha_{t}} \bx_{t}, \quad \tilde \beta_{t} \coloneqq \frac{1-\bar \alpha_{t-1}}{1 - \bar \alpha_{t}} \beta_{t}
\end{align}

where $\alpha_t^{\text{DDPM}} \coloneqq 1-\beta_t$ and $\bar \alpha_t \coloneqq \prod_{s=1}^t \alpha_s^{\text{DDPM}}$.
DDPM typically assumes that $q(\bx_{t-1}\mid\bx_t,\bx_0)$ and $p_\theta(\bx_{t-1}\mid\bx_t)$ share the same noise scale, i.e., $\boldsymbol{\Sigma}_{\theta}(\bx_t,t)=\tilde \beta_t$.
Under this assumption, $L_{t-1}$ (the KL divergence between $q(\bx_{t-1}\mid\bx_t,\bx_0)$ and $p_\theta(\bx_{t-1}\mid\bx_t)$) can be expressed by directly comparing the corresponding means $\boldsymbol{\tilde \mu}_t$ and $\boldsymbol{\mu}_\theta$:

\begin{align}
    \label{eq:ELBO_mu}
    L_{t-1}=\mathbb{E}_q\left[\frac{1}{2\sigma_t^2}\|\boldsymbol{\tilde \mu}_t(\bx_t, \bx_0)-\boldsymbol{\mu}_\theta(\bx_t, t)\|^2\right] + C
\end{align}

where $C$ is a constant value. $\boldsymbol{\mu}_\theta(\bx_t, t)$ can be represented with noise-prediction model $\boldsymbol{\epsilon}_\theta$:

\begin{align}
    \label{eq:mu_epsilon}
    \boldsymbol{\mu}_\theta(\bx_t, t) = \boldsymbol{\tilde \mu}\left(\bx_t, \frac{1}{\sqrt{\bar \alpha_t^\text{DDPM}}} \left(\bx_t-\sqrt{1-\bar \alpha_t^{\text{DDPM}}}\boldsymbol{\epsilon}_\theta(\bx_t) \right) \right) = \frac{1}{\sqrt{\alpha_t^{\text{DDPM}}}}\left(\bx_t-\frac{\beta_t}{\sqrt{1- \bar \alpha_t^{\text{DDPM}}}}\boldsymbol{\epsilon}_\theta(\bx_t, t) \right)
\end{align}

For a detailed derivation, we refer the reader to the original paper~\cite{ho2020denoising}.
Notably, $\boldsymbol{\mu}_\theta$ can be expressed as a function of $\boldsymbol{\tilde \mu}_t$, which motivates the implementations of $\texttt{CalcMean\_Q}$ and $\texttt{CalcMeanStd\_P}$ used in our codec.

We provide pseudocode for $\texttt{CalcMean\_Q}$ and $\texttt{CalcMeanStd\_P}$ in Algorithms~\ref{alg:calc_mean_q} and~\ref{alg:calc_mean_std_p}, respectively.
Given noise schedule parameters $\alpha_t,\alpha_s,\sigma_t,\sigma_s$ (with $s\leq t$), $\bx_0$ for $\texttt{CalcMeanStd\_Q}$, and $\boldsymbol{\epsilon}_\theta$ for $\texttt{CalcMeanStd\_P}$, each function computes the required quantities based on Eqs.~\ref{eq:mu_forward_closed_form} and~\ref{eq:mu_epsilon}.
In DiffC, the noise schedule parameters are $\alpha_t=\sqrt{\bar \alpha_t^{\text{DDPM}}}$ and $\sigma_t=\sqrt{1-\bar \alpha_t^{\text{DDPM}}}$.

\begin{figure*}[t]
\begin{minipage}[t]{0.5\textwidth}
\begin{algorithm}[H]
 \caption{$\texttt{CalcMean\_Q}$} \label{alg:calc_mean_q}
 \small
 \begin{algorithmic}[1]
    \REQUIRE $\bx_t, \bx_0, \alpha_t, \sigma_t, \alpha_s, \sigma_s$
    \ENSURE $\boldsymbol{\mu}_t$
    \STATE $\boldsymbol{\mu}_t = \frac{\alpha_s(1-\alpha_t)}{\sigma_t^2}\bx_0 + \frac{\frac{\alpha_t}{\alpha_s}\sigma_s^2}{\sigma_t^2}\bx_t$
    \STATE Return $\boldsymbol{\mu}_t$
    \STATE \rule{0pt}{3.25ex} no operation
 \end{algorithmic}
\end{algorithm}
\end{minipage}
\hfill
\begin{minipage}[t]{0.5\textwidth}
\begin{algorithm}[H]
 \caption{$\texttt{CalcMeanStd\_P}$} \label{alg:calc_mean_std_p}
 \small
 \begin{algorithmic}[1]
    \REQUIRE $\bx_t, \boldsymbol{\epsilon}_\theta, \alpha_t, \sigma_t, \alpha_s, \sigma_s$ 
    \ENSURE $\boldsymbol{\mu}_\theta, \quad \tilde \beta_t$
    \STATE $\boldsymbol{\mu}_\theta = \frac{\alpha_s(1-\alpha_t)}{\sigma_t^2}\boldsymbol{\bx}_\theta + \frac{\frac{\alpha_t}{\alpha_s}\sigma_s^2}{\sigma_t^2}\bx_t$
    \STATE $\tilde \beta_t = \frac{1-\bar \alpha_s}{1-\bar \alpha_t}\beta_t$
    \STATE Return $\boldsymbol{\mu}_\theta, \quad \tilde \beta_t$
 \end{algorithmic}
\end{algorithm}
\end{minipage}
\vspace{-1em}
\end{figure*}

\section{Velocity--Noise Parameterization Equivalence for Flow Models}

In the continuous-time setting, the forward process is described by the SDE

\begin{equation}
   {d}\bx_t
=
\boldsymbol{f}(\bx_t, t)\,dt
+
g(t)\,d\boldsymbol{w}_t,
\end{equation}

where $\boldsymbol{f}(\bx_t,t)$ is the drift coefficient, $g(t)$ is the diffusion coefficient, and $\boldsymbol{w}_t$ is a standard Wiener process.
This SDE models gradual Gaussian perturbations over time, such that the distribution of $\bx_t$ approaches a simple prior (e.g., a standard Gaussian) as $t$ increases.

The reverse-time process is obtained by solving the corresponding reverse-time SDE:

\begin{equation}
\label{eq:reverse-SDE}
d\bx_t
=
\Bigl[
\boldsymbol{f}(\bx_t, t)
-
g(t)^2 \nabla_{\bx} \log p_t(\bx)
\Bigr]dt
+
g(t)\,d\bar{\boldsymbol{w}}_t
\end{equation}

where $\bar{\boldsymbol{w}}_t$ is a standard Wiener process in reverse time, $p_t(\bx)$ is the marginal density of $\bx_t$, and $\nabla_{\bx}\log p_t(\bx)$ is the score function, which is the estimation target in diffusion models~\cite{song2020score}.
It is well known that the score can be parameterized via a noise-prediction model $\boldsymbol{\epsilon}_\theta$:

\begin{equation}
\label{eq:score_epsilon}
\nabla_{\boldsymbol{\bx}} \log p_t(\boldsymbol{\bx})
=
-\frac{1}{\sigma_t}\,
\boldsymbol{\epsilon}_\theta(\bx_t, t)
\end{equation}

Direct simulation of Eq.~\ref{eq:reverse-SDE} can be unstable due to the stochastic term.
To obtain a deterministic alternative, Song et al.~\cite{song2020score} proposed the probability flow ODE (PF-ODE):

\begin{equation}
\label{eq:PF-ODE}
\frac{d\bx_t}{dt}
=
\boldsymbol{f}(\bx_t, t)
-
\frac{1}{2} g(t)^2 \nabla_{\bx} \log p_t(\bx).
\end{equation}

The PF-ODE induces the same marginal distributions $p_t(\bx)$ as the reverse-time SDE, while enabling stable generation with deterministic numerical solvers.

Following the velocity--noise parameterization equivalence discussed by Lai et al.~\cite{lai2025principles}, we summarize here the specialization used in this paper. Using Eq.~\ref{eq:score_epsilon} and the noise schedule parameters $\alpha_t$ and $\sigma_t$, the PF-ODE in Eq.~\ref{eq:PF-ODE} can be rewritten with $\boldsymbol{\epsilon}_\theta$ as follows:

\begin{align}
    \label{Rectified_Flow_Reverse}
    \frac{d\bx_t}{dt} &= \frac{\alpha_t'}{\alpha_t}\bx_t - \frac{1}{2}\alpha_t^2\frac{d}{dt}(\frac{\sigma_t^2}{\alpha_t^2})\nabla_{\bx_t}\log p_t(\bx_t) \\
    &= \frac{\alpha_t'}{\alpha_t}\bx_t + \frac{1}{2}\alpha_t^2\frac{d}{dt}(\frac{\sigma_t^2}{\alpha_t^2})\frac{\boldsymbol{\epsilon}_\theta(\bx_t,t)}{\sigma_t} \\
    &= \frac{\alpha_t'}{\alpha_t}\bx_t + \frac{1}{2}\alpha_t^2 \frac{2\sigma_t\sigma_t'\alpha_t^2-2\alpha_t\alpha_t'\sigma_t^2}{\alpha_t^4}\frac{\boldsymbol{\epsilon}_\theta(\bx_t,t)}{\sigma_t} \\
    &= \frac{\alpha_t'}{\alpha_t}\bx_t + (\sigma_t' - \frac{\alpha_t'\sigma_t}{\alpha_t})\boldsymbol{\epsilon}_\theta(\bx_t,t)
\end{align}

Here, the velocity field $\boldsymbol{v}_\theta$ approximates the flow of data between two distributions and a new sample can be generated by solving an ODE:

\begin{equation}
    \label{velocity_ode}
    \frac{d \bx_t}{d t} = \bv_\theta(\bx_t, t)
\end{equation}

Comparing Eq.~\ref{Rectified_Flow_Reverse} and \ref{velocity_ode}, we obtain

\begin{align}
\boldsymbol{v}_\theta(\bx_t, t) &= \frac{\alpha_t'}{\alpha_t}\bx_t + (\sigma_t' - \frac{\alpha_t'\sigma_t}{\alpha_t})\boldsymbol{\epsilon}_\theta(\bx_t,t) \\
\boldsymbol{\epsilon}_\theta(\bx_t,t) &= \frac{\boldsymbol{v}_\theta(\bx_t,t)-\frac{\alpha_t'}{\alpha_t}\bx_t}{\sigma_t'-\frac{\alpha_t'\sigma_t}{\alpha_t}}
\label{eq:v_eps_relation}
\end{align}

which corresponds directly to Eq.~\ref{RF_velocity_to_epsilon} when $\boldsymbol{v}_\theta=\boldsymbol{v}_\theta^\text{RF}, \alpha_t=\alpha_\text{RF}(t), \sigma_t=\sigma_\text{RF}(t)$ and Eq.~\ref{MF_velocity_to_epsilon} when $\boldsymbol{v}_\theta=\boldsymbol{v}_\theta^\text{MF}, \alpha_t=\alpha_\text{MF}(t), \sigma_t=\sigma_\text{MF}(t)$.

\section{Details of DiffC (RF)}
\label{app:diffc-rf}
To apply Rectified Flow models (e.g., Flux~\cite{labs2025flux1kontextflowmatching}) within the DiffC framework, Vonderfecht et al.~\cite{vonderfecht2025lossy} aligned the Rectified Flow noise schedule to that of DDPM.
Specifically, they performed the following two operations:

1. \textbf{Adjusting the Time Steps}: Map DDPM time steps to corresponding Rectified-Flow time steps so that the two processes match the signal-to-noise ratio at time $t$:

\begin{align}
    \label{eq:ddpm_to_flow_time_steps}
    \frac{\alpha_\text{RF}(t)}{\sigma_\text{RF}(t)} = \frac{\sqrt{\bar \alpha_t^{\text{DDPM}}}}{1-\sqrt{\bar \alpha_t^{\text{DDPM}}}}
\end{align}

2. \textbf{Scaling $\bx_t$}: Since $\bx_t$ is scaled for DDPM denoisers, it is rescaled to match the input scale expected by the Rectified-Flow velocity network using the scalar $c$ defined as follows:

\begin{align}
    \label{eq:scale_for_ddpm_to_rf_translation}
    c = \frac{\sqrt{\bar \alpha_t^{\text{DDPM}}}}{\alpha_\text{RF}(t)}, \quad \bx_t^\text{scaled} = \frac{\bx_t}{c}
\end{align}

When applying Rectified Flow within DiffC, the noise schedule parameters and $\bx_t$ are translated accordingly, and the scaled latent $\bx_t^{\text{scaled}}$ is inputted to the Rectified Flow model.
Although DiffC (RF) achieves competitive performance compared to existing methods, its denoising quality remains suboptimal, as demonstrated in results sections.

\begin{figure*}
 \centering
 \includegraphics[width=\textwidth]{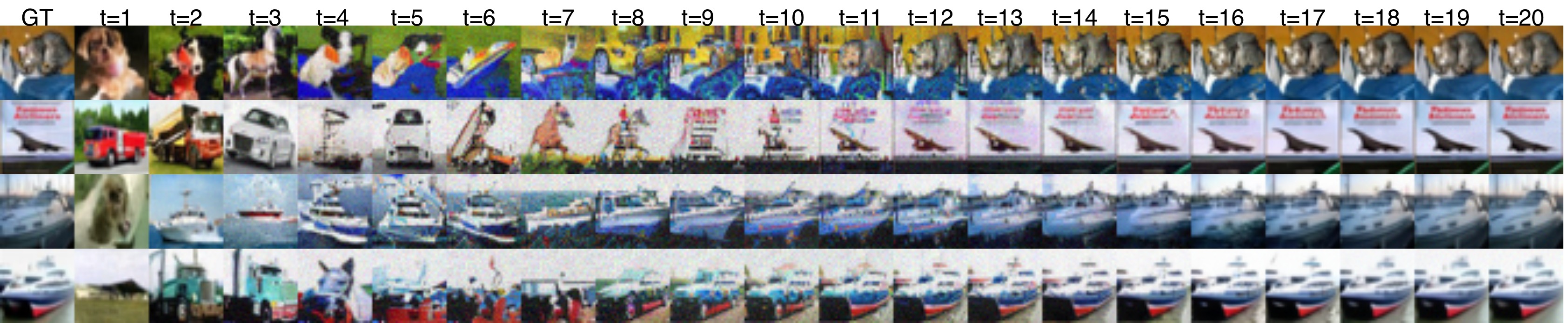}
 \caption{\label{fig:mefc_steps_cifar10} Visualization of MeFC reconstructions on CIFAR10 from different time steps $t$. For small $t$ (up to $t=3$), reconstructions may differ from the input but remain perceptually plausible; for larger $t$, the decoder increasingly reconstructs the input with added Gaussian noise, which is then gradually removed.}
 \end{figure*}

\section{Results}

\subsection{Public Code Repositories}
\label{sec:repositories}

Table~\ref{table:repositories} lists the publicly available web cite, GitHub and Hugging Face repositories of the datasets and generative models used in our experiments.
All implementations and pre-trained models were obtained from these sources without modification unless otherwise stated.

\begin{table}[h]
\centering
\caption{Datasets and publicly available GitHub and Hugging Face repositories used in our experiments.}
\label{table:repositories}
\begin{tabular}{ll}
\toprule
Name & Repository \\
\midrule
ImageNet 64$\times$64 &
\href{https://huggingface.co/datasets/benjamin-paine/imagenet-1k-64x64}{benjamin-paine/imagenet-1k-64x64} \\
ImageNet 256$\times$256 &
\href{https://huggingface.co/datasets/benjamin-paine/imagenet-1k-256x256}{benjamin-paine/imagenet-1k-256x256} \\
Improved DDPM (for 32$\times$32) &
\href{https://github.com/openai/improved-diffusion}{github.com/openai/improved-diffusion} \\
Guided Diffusion (for 64$\times$64, 256$\times$256) &
\href{https://github.com/openai/guided-diffusion}
{github.com/openai/guided-diffusion} \\
Rectified Flow &
\href{https://github.com/sangyun884/rfpp}{github.com/sangyun884/rfpp} \\
Flux &
\href{https://huggingface.co/docs/diffusers/main/api/models/flux_transformer}{huggingface.co/docs/diffusers/main/api/models/flux\_transformer} \\
CTM &
\href{https://github.com/sony/ctm}{github.com/sony/ctm} \\
CTM 32$\times$32 &
\href{https://github.com/Kim-Dongjun/ctm-cifar10}{github.com/Kim-Dongjun/ctm-cifar10} \\
MeanFlow &
\href{https://github.com/zhuyu-cs/MeanFlow}{github.com/zhuyu-cs/MeanFlow} \\
\bottomrule
\end{tabular}
\end{table}

\subsection{Supported Image Resolutions}
\label{sec:resolution_support}

Table~\ref{table:resolution_support} summarizes the image resolutions supported by each few-step generative model used in our experiments.
A checkmark indicates that a publicly available pre-trained model exists and was used for evaluation at the corresponding resolution.
A dash denotes that no such pre-trained model was publicly available at the time of evaluation, and the corresponding experiment was therefore omitted.

\begin{table}[h]
\centering
\caption{Supported image resolutions for each model.}
\label{table:resolution_support}
\begin{tabular}{lccc}
\toprule
Model & $32\times32$ & $64\times64$ & $256\times256$ \\
\midrule
Rectified Flow  & \checkmark & \checkmark & \checkmark \\
CTM             & \checkmark & \checkmark & --          \\
MeanFlow        & \checkmark & --          & \checkmark  \\
\bottomrule
\end{tabular}
\end{table}

\subsection{Performance Evaluation with Varying Forward/Reconstruction Steps}
\subsubsection{Varying Forward Steps}
\begin{figure*}[t]
\centering
\includegraphics[width=\textwidth]{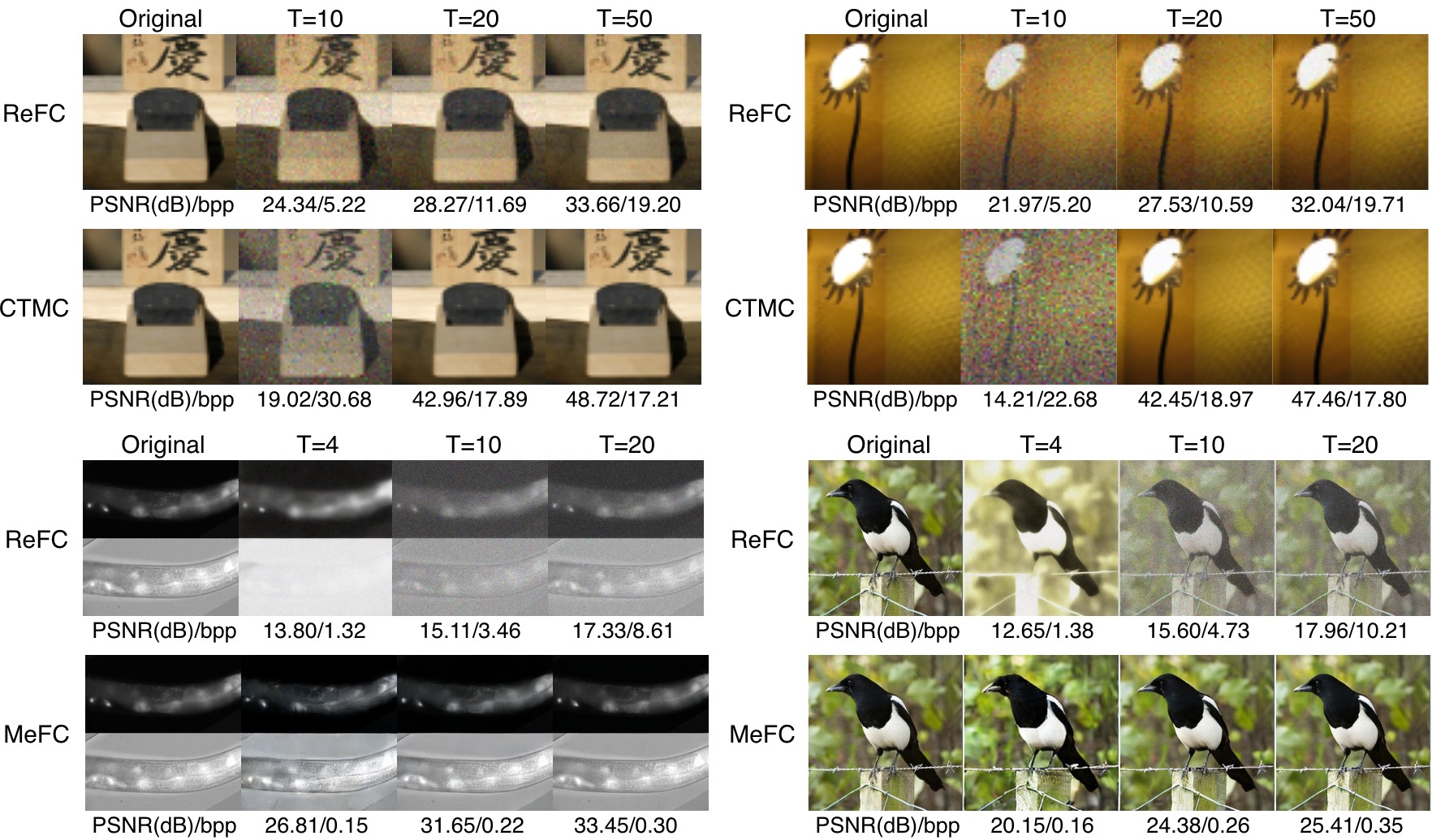}
\caption{\label{fig:steps_comparison_imagenet64_256} Reconstructed image comparison by changing the number of forward steps $T$. The top two rows show results for ImageNet 64$\times$64, while the bottom two rows show results for ImageNet 256$\times$256, which have been scaled to 64$\times$64 resolutions for easier viewing.}
\end{figure*}

We vary the number of forward steps $T$ for ReFC and CTMC on ImageNet $64\times64$ and for ReFC and MeFC on ImageNet $256\times256$.
For each dataset, we measure average encoding and decoding time over 10 images and report the results in Table~\ref{tab:time_refc_mefc}; qualitative reconstructions are shown in Fig.~\ref{fig:steps_comparison_imagenet64_256}.

Increasing $T$ generally increases both encoding and decoding time.
However, CTMC does not necessarily exhibit a monotonic increase in encoding time with larger $T$.
This behavior is influenced by the RCC implementation, which explicitly estimates the KL divergence between the sender distribution $q$ and the shared distribution $p$ and can become slower when the divergence is large (see~\cite{vonderfecht2025lossy} for implementation details).
In DiffC, the reverse process is trained to match the forward posteriors and the means of $q$ and $p$ can be computed analytically, which stabilizes the KL divergence and typically avoids slowdowns.
In contrast, CTMC relies on an assumed Gaussian family centered at an $\bx_0$ estimate, and minimizing the KL divergence is not guaranteed.
Empirically, larger $T$ can improve the accuracy of $\bx_0$ estimation under $p$, which stabilizes the KL divergence and can reduce RCC overhead.

Across methods, larger $T$ improves reconstruction quality (PSNR), as shown in Fig.~\ref{fig:steps_comparison_imagenet64_256}.
Considering the speed-quality trade-off, we adopt the configurations in Tables~\ref{table:ImageNet 64x64} and~\ref{table:ImageNet 256x256}.

\begin{table}[t]
\centering
\caption{Encoding and decoding time comparison under different forward steps $T$. (a) shows the results of ReFC and CTMC on ImageNet 64$\times$64, while (b) summarizes the results of ReFC and MeFC on ImageNet 256$\times$256.}
\label{tab:time_refc_mefc}
\begin{minipage}{0.8\linewidth}
\centering
\caption*{(a) ImageNet 64$\times$64}
\begin{tabular}{l c c c}
\toprule
Method & $T$ & Encoding Time (s) & Decoding Time (s) \\
\midrule
\multirow{2}{*}{ReFC} & 10  & 2.70 & 1.90 \\
     & 50 & 10.94 & 47.49 \\
\multirow{2}{*}{CTMC} & 10  & 14.41 & 0.57 \\
& 50 & 11.08 & 4.40 \\
\bottomrule
\end{tabular}
\end{minipage}
\hfill
\begin{minipage}{0.8\linewidth}
\centering
\caption*{(b) ImageNet 256$\times$256}
\begin{tabular}{l c c c}
\toprule
Method & $T$ & Encoding Time (s) & Decoding Time (s) \\
\midrule
\multirow{2}{*}{ReFC} & 4  & 13.18 & 3.06 \\
     & 10 & 38.46 & 22.53 \\
\multirow{2}{*}{MeFC} & 4  & 0.94 & 0.42 \\
& 20 & 2.36 & 5.68 \\
\bottomrule
\end{tabular}
\end{minipage}
\end{table}

\subsubsection{Varying Reconstruction Steps}

\begin{table}
\centering
\caption{Encoding and decoding time comparison under different reverse process steps (Rev Steps). (a) shows the results of ReFC and CTMC on ImageNet 64$\times$64, and (b) summarizes the results of ReFC and MeFC on ImageNet 256$\times$256.}
\label{tab:dec_time_refc_mefc}
\begin{minipage}{0.8\linewidth}
\centering
\caption*{(a) ImageNet 64$\times$64}
\begin{tabular}{l c c}
\toprule
Method & Rev Steps & Decoding Time (s) \\
\midrule
\multirow{2}{*}{ReFC} & 1  & 1.48 \\
     & 4 & 2.73 \\
\multirow{2}{*}{CTMC} & 4  & 2.72 \\
& 20 & 3.28 \\
\bottomrule
\end{tabular}
\end{minipage}
\hfill
\begin{minipage}{0.8\linewidth}
\centering
\caption*{(b) ImageNet 256$\times$256}
\begin{tabular}{l c c}
\toprule
Method & Rev Steps & Decoding Time (s) \\
\midrule
\multirow{2}{*}{ReFC} & 1  & 10.80 \\
     & 4 & 27.73 \\
\multirow{2}{*}{MeFC} & 1  & 1.07 \\
& 4 & 1.57 \\
\bottomrule
\end{tabular}
\end{minipage}
\end{table}

We study how the number of reconstruction steps affects both visual quality and runtime. Table~\ref{tab:dec_time_refc_mefc} summarizes decoding time for different reconstruction step budgets.
A step count of 1 corresponds to one-step reconstruction from time $t$.
For 4-step reconstruction, we solve the ODE from $t$ using Euler method over four steps (or $t$ steps when $t<4$).
As expected, fewer steps reduce decoding time; however, few-step reconstruction, including one-step, can introduce noticeable artifacts.
We also observe that Euler-based reconstruction can exhibit abrupt changes (e.g., around $t=8$), whereas the denoising-based reconstruction removes noise more gradually.
This suggests that denoising-based reconstruction is preferable in the low-bit regime.
Designing Rectified-Flow-based codecs that achieve both high speed and strong low-bit performance, ideally in a single step, remains an important direction for future work.

\begin{figure*}[p]
\centering
\includegraphics[width=\textwidth]{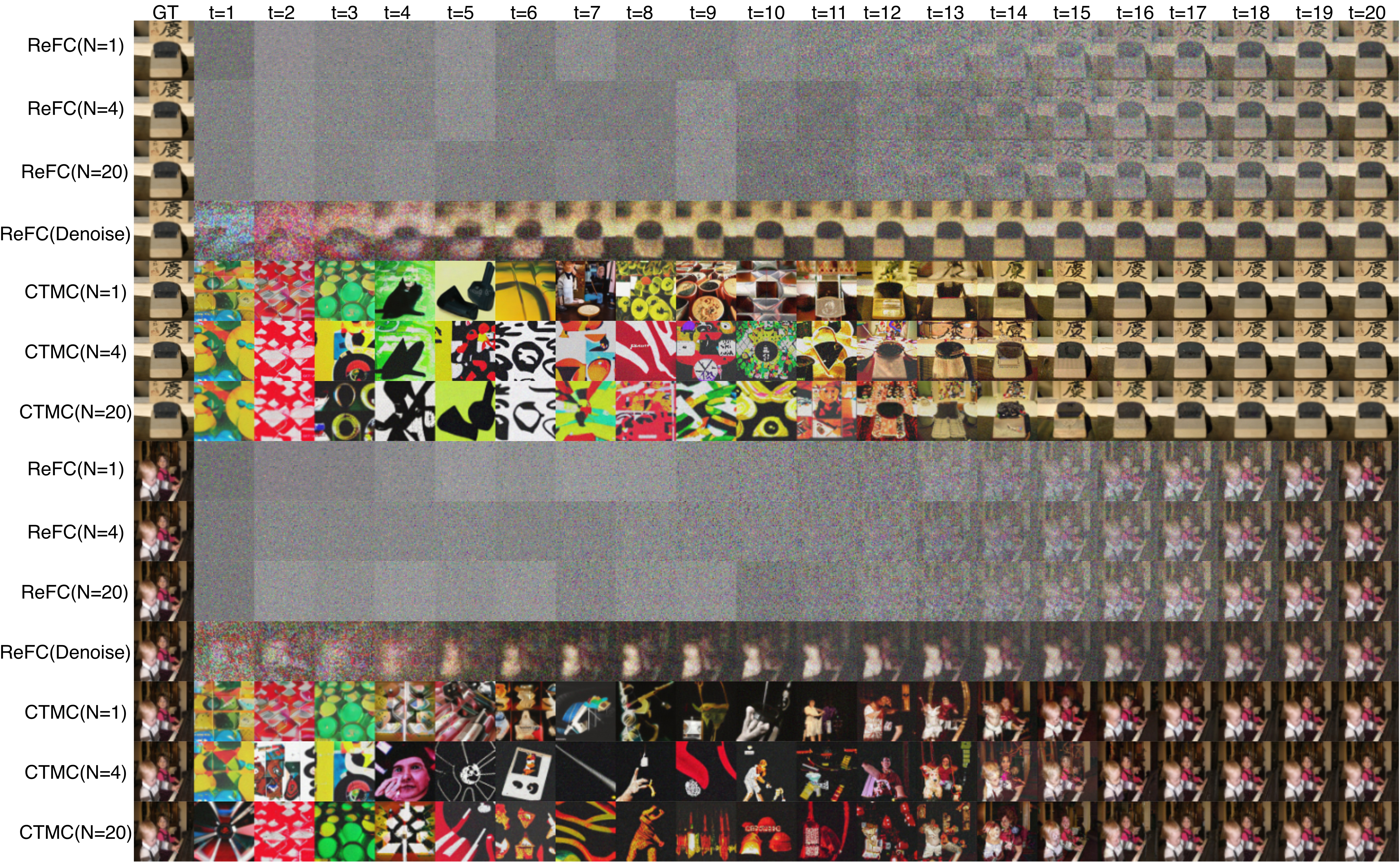}
\caption{\label{fig:rev_steps_comparison_imagenet64} Visual comparison results of varying the reconstructed methods for ReFC and MeFC on ImageNet 64$\times$64. The displayed results show the reconstructed images from each timestep $t$ over $N$ steps. For ReFC, Euler method was applied with $N=1,4,20$. When $t \leq N$, the reconstruction performed over $t$ steps. ReFC (Denoise) refers to a reconstruction method that employs $\boldsymbol{\epsilon}$-prediction with $\boldsymbol{\epsilon}_\theta^\text{RF}$ such like DDPM. CTMC also applies Euler method with $N=4,20$, as well as direct image estimation from $t$ by $\bx_\theta^\text{CTM}$ when N$=$1.}
\end{figure*}

\begin{figure*}[p]
\centering
\includegraphics[width=\textwidth]{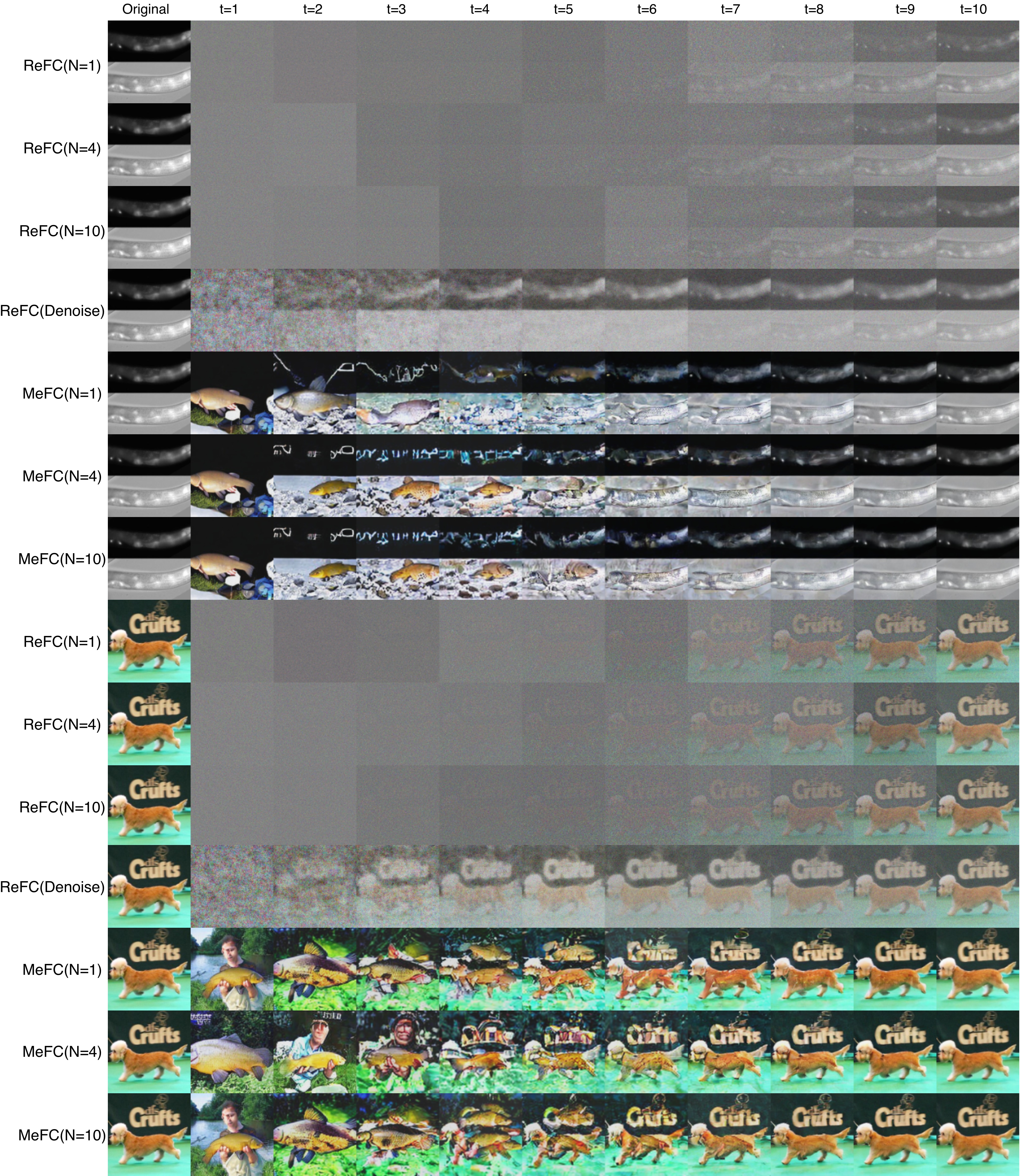}
\caption{\label{fig:rev_steps_comparison_imagenet256} Reconstructed image comparison by changing the reconstruction steps on ImageNet 256$\times$256. For MeFC, Euler method is applied for $N=4, 10$ and in the case of $N=1$, average velocity field $\boldsymbol{u}_\theta^\text{MF}$ is used for one-step generation such as MeanFlow. ReFC applies Euler method when $N=1,4,10$ and $\boldsymbol{\epsilon}$-prediction (ReFC (Denoise)).}
\end{figure*}

\subsection{ImageNet 64$\times$64}
\begin{figure*}
 \centering
 \includegraphics[width=\textwidth]{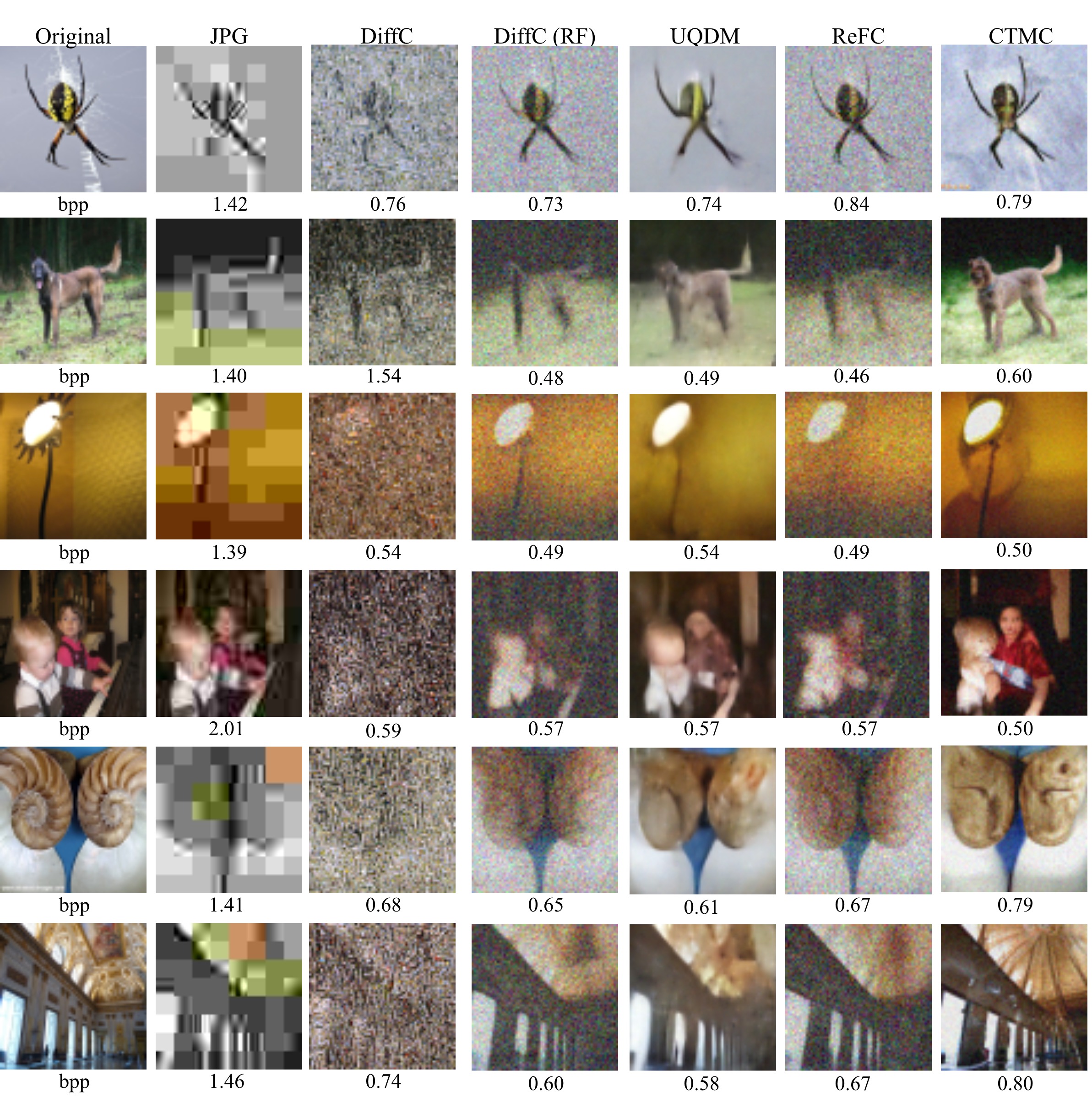}
 \caption{\label{fig:ImageNet64_reconstructed_image_comparison} Visual Comparison of each image compression method including JPEG, DiffC, DiffC (RF), UQDM, ReFC and CTMC, chosen at roughly same bpp.}
 \end{figure*}

Fig.~\ref{fig:ImageNet64_reconstructed_image_comparison} shows a qualitative comparison on ImageNet $64\times$64 among DiffC, DiffC (RF), UQDM, ReFC, and CTMC.
JPEG often misses colors in some regions, and DiffC can produce noisy reconstructions due to the limited quality of the underlying DDPM backbone at this resolution.
When comparing UQDM and CTMC at the same bit rate, UQDM outputs smooth images, while CTMC restores fine details, resulting in clearly defined objects. From these results, CTMC reconstructs images with higher fidelity.

\subsection{ImageNet 256$\times$256}

\begin{table}[h]
\centering
\caption{Comparison of the number of steps for both forward ("Fwd") and reverse ("Rev"), and average encoding (Enc) and decoding (Dec) times in seconds. All results were measured on a single RTX 3090 GPU for ImageNet 256$\times$256.}
\setlength{\tabcolsep}{2.1mm}
\label{table:ImageNet 256x256}
\begin{tabular}{cccccc}
\cline{1-5}
Method & Fwd Steps & Rev Steps & Enc (s) & Dec (s) &  \\
\cline{1-5}
DiffC  & 50 & 50     &    206.46    & 102.32 &  \\
DiffC (RF) & 20 & 20     &    88.61    & 93.75 &  \\
PerCo  & - & -  &   0.80    &   0.57    &  \\
ReFC & 20 & 20   &  84.93  & 93.04  &  \\
MeFC   & 10 &  10   &   2.35     &     1.84    &  \\
\cline{1-5}
\end{tabular}
\end{table}
 
We compared the performance of DiffC, DiffC (RF), PerCo, MeFC, and ReFC on ImageNet 256$\times$256.
Table~\ref{table:ImageNet 256x256} summarizes the configuration of each method, including the numbers of forward and reverse steps, and the encoding and decoding times in seconds. These results indicate that MeFC substantially reduces encoding and decoding time relative to DiffC. Although PerCo remains faster in both encoding and decoding, MeFC appears to offer a favorable quality--speed trade-off on this benchmark.

\begin{figure*}
 \centering
 \includegraphics[width=\textwidth]{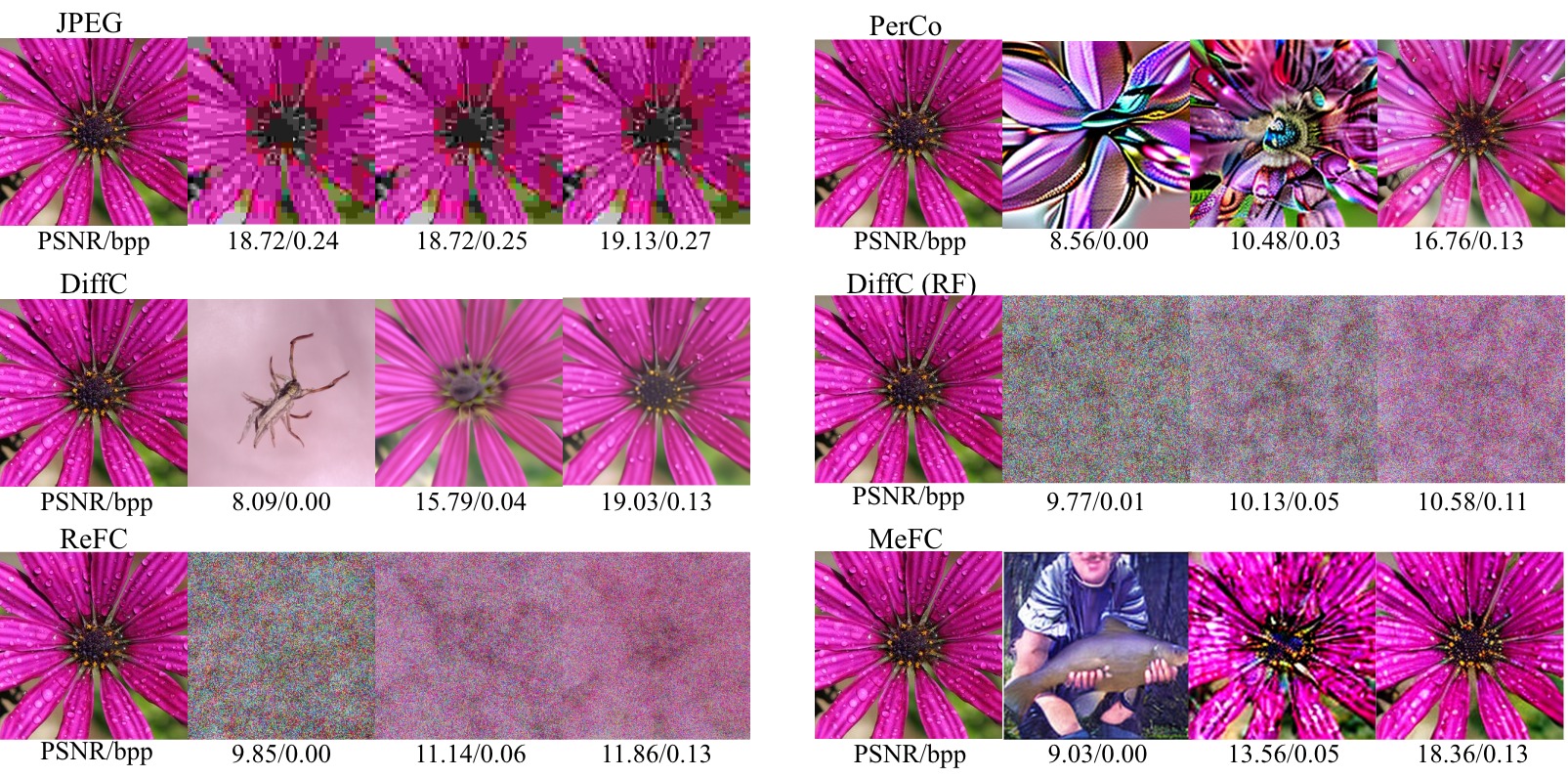}
 \caption{\label{fig:ImageNet256_visual_comparison} Visual Comparison of image compression methods including JPEG, PerCo, DiffC, DiffC (RF), ReFC and MeFC on ImageNet 256$\times$256.
.}
 \end{figure*}

Fig.~\ref{fig:ImageNet256_visual_comparison} compares reconstructions across bit rates, which suggests that MeFC produces higher fidelity images than PerCo in the low-bit-rate regime.

\begin{figure*}
 \centering
 \includegraphics[width=\textwidth]{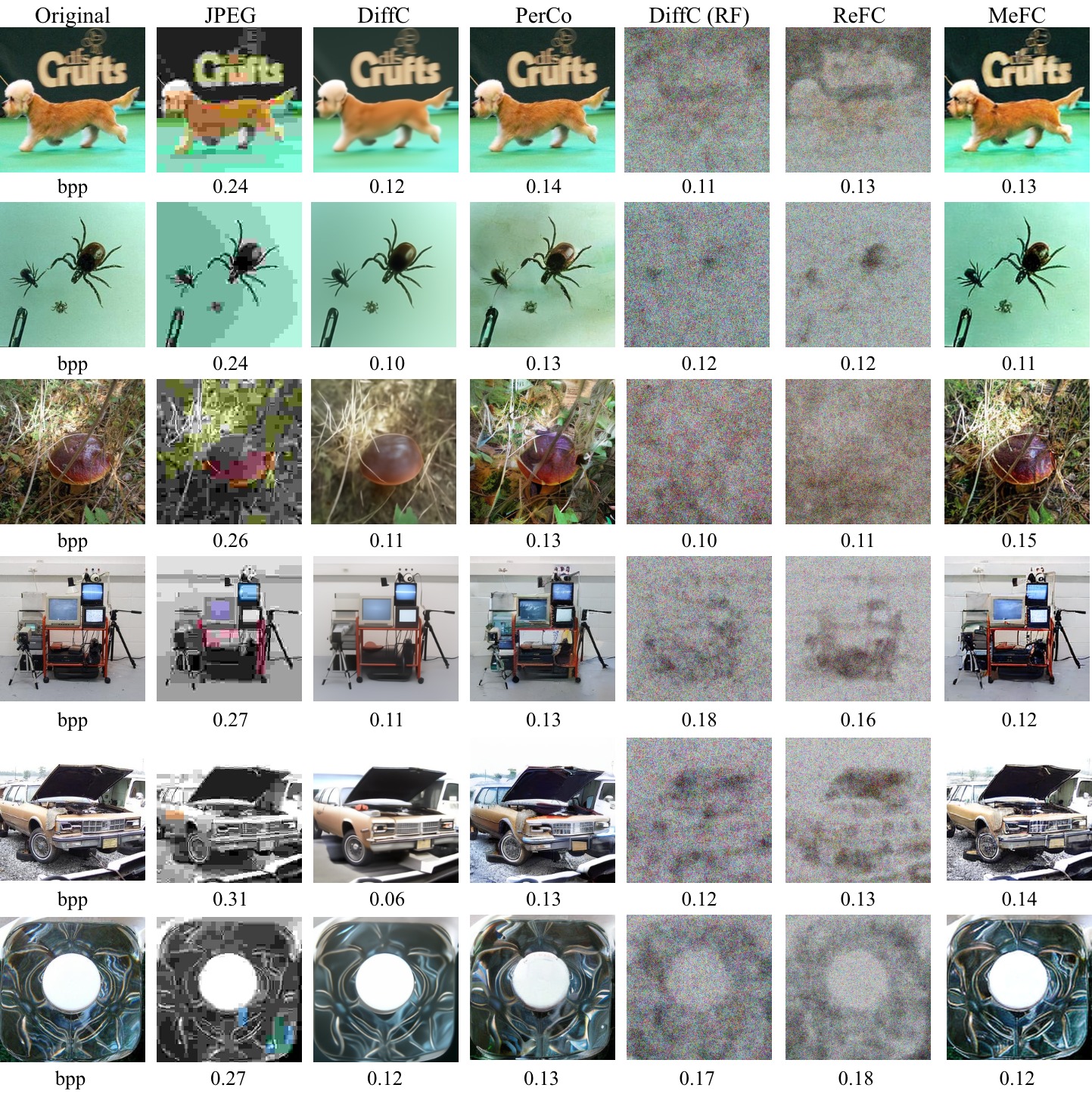}
 \caption{\label{fig:ImageNet256_reconstructed_image_comparison} Visual Comparison of reconstructed images on ImageNet 256$\times$256. We roughly choose the similar bpp.
.}
 \end{figure*}

We report an additional visual comparison of reconstructed images in Fig.~\ref{fig:ImageNet256_reconstructed_image_comparison}.
Consistent with the observations above, MeFC tends to produce more natural reconstructions than the competing methods in the low-bit-rate regime, while ReFC and DiffC (RF) exhibit similar tendencies. ReFC nevertheless captures objects within the image more distinctly, consistent with the results on the other datasets. Expressing the model in the denoising parameterization via its learned velocity field appears to enable more effective noise removal in the reconstructed images.


% You can have as much text here as you want. The main body must be at most $8$
% pages long. For the final version, one more page can be added. If you want, you
% can use an appendix like this one.

% The $\mathtt{\backslash onecolumn}$ command above can be kept in place if you
% prefer a one-column appendix, or can be removed if you prefer a two-column
% appendix.  Apart from this possible change, the style (font size, spacing,
% margins, page numbering, etc.) should be kept the same as the main body.
%%%%%%%%%%%%%%%%%%%%%%%%%%%%%%%%%%%%%%%%%%%%%%%%%%%%%%%%%%%%%%%%%%%%%%%%%%%%%%%
%%%%%%%%%%%%%%%%%%%%%%%%%%%%%%%%%%%%%%%%%%%%%%%%%%%%%%%%%%%%%%%%%%%%%%%%%%%%%%%

\end{document}